\documentclass[times,twocolumn,final,authoryear]{elsarticle}

\usepackage{ycviu}
\usepackage{framed,multirow}

\usepackage{amsmath, amssymb} 
\usepackage{latexsym}
\usepackage{colortbl}
\usepackage{makecell}
\usepackage{booktabs}

\usepackage{url}
\usepackage{xcolor}
\definecolor{newcolor}{rgb}{.8,.349,.1}
\usepackage{setspace}

\journal{Computer Vision and Image Understanding}

\begin{document}
\thispagestyle{empty}

\ifpreprint
  \setcounter{page}{1}
\else
  \setcounter{page}{1}
\fi

\begin{frontmatter}

\title{View Consistency Aware Holistic Triangulation for 3D Human Pose Estimation}

\author[1]{Xiaoyue \snm{Wan}} 
\author[1]{Zhuo \snm{Chen}}
\author[1]{Xu \snm{Zhao}\corref{cor1}}
\cortext[cor1]{Corresponding author: }
\ead{zhaoxu@sjtu.edu.cn}

\address[1]{School of Electronic Information and Electrical Engineering, Shanghai Jiao Tong University, Shanghai, 200000, China}

\received{1 May 2013}
\finalform{10 May 2013}
\accepted{13 May 2013}
\availableonline{15 May 2013}
\communicated{S. Sarkar}

\begin{abstract}
The rapid development of multi-view 3D human pose estimation (HPE) is attributed to the maturation of monocular 2D HPE and the geometry of 3D reconstruction. However, 2D detection outliers in occluded views due to neglect of view consistency, and 3D implausible poses due to lack of pose coherence, remain challenges. 
To solve this, we introduce a Multi-View Fusion module to refine 2D results by establishing view correlations.
Then, Holistic Triangulation is proposed to infer the whole pose as an entirety, and anatomy prior is injected to maintain the pose coherence and improve the plausibility. Anatomy prior is extracted by PCA whose input is skeletal structure features, which can factor out global context and joint-by-joint relationship from abstract to concrete. Benefiting from the closed-form solution, the whole framework is trained end-to-end.
% Extensive experiments demonstrate that MVF is capable of enhancing views consistency and our anatomy prior can maintain pose coherence.
Our method outperforms the state of the art in both precision and plausibility which is assessed by a new metric.
\end{abstract}


\end{frontmatter}

%\linenumbers

\section{Introduction}\label{}
% background
3D human pose estimation (HPE) is a significant computer vision problem with numerical applications such as human behavior analysis, X-reality, etc~\citep{WANG2021103225}. 
To estimate 3D pose, there are two sensor setting streams: monocular~\citep{martinez2017simple,pavlakos2017coarse,xu2021graph} and multi-view~\citep{gavrila19963,burenius20133d}. In this paper, we focus on 
multi-view 3D HPE, for its capability to estimate absolute 3D position without inherent depth ambiguities which monocular suffers.

% problem
One of the most common frameworks~\citep{iskakov2019learnable,dong2019fast,remelli2020lightweight,kocabas2019self} of multi-view methods follows a two-step procedure: (1) detect 2D keypoints of human skeleton at each view separately, (2) apply Linear Triangulation (LT) which utilizes epipolar geometry~\citep{2003Multiple} to reconstruct 3D pose. The framework is elegant because 2D detectors can be off-the-shelf and closed-form solution LT enables end-to-end training but without any learning cost. However, there are still two main drawbacks: (1) 2D keypoints detected in each view are independent of each other, and will be hampered by the occlusion and overlap due to lack of view consistency. (2) LT in step 2 calculates each 3D joint individually, neglecting the global context of whole pose. Hence, it is unable to identify the 3D outliers, which usually causes implausible poses. 
\vspace{-0.15ex}

% method
To solve the first problem, Multi-View Fusion (MVF) module is proposed to refine the 2D keypoint by establishing view correlations. We argue that multiple image points projected from a 3D point share similar representations. In another word, two most similar points in different views are mostly intersected to one 3D point. According to this assumption, MVF utilizes keypoints detected in source views to generate pseudo heatmaps which represents the probability distribution the keypoint localized in reference view through feature matching. And pseudo heatmaps can guide the reference keypoints to perceive other views. There are also some works aimed to enhance view consistency through feature fusion: the fully-connected CrossView~\citep{qiu2019cross} and the epipolar sample fusion in Epipolar Transformer~\citep{he2020epipolar}. But, in MVF, the utilization of the detected keypoint location makes calculation more efficient, and the pseudo heatmap guidance is also more intuitive than feature fusion.
% all features in reference view still need to be aggregated because ignoring the keypoint location information. Our MVF leverages the initial 2D localization to sample joint feature and generates the corresponding pseudo heatmaps which represent the probability distribution the keypoint localized in other views. Then, heatmaps corresponding to different views are aggregated. Thus, 2D outliers could be eliminated through perceiving all views.
\vspace{-0.15ex}

\begin{figure*}[!t]
\centering
\includegraphics[width=0.8\textwidth]{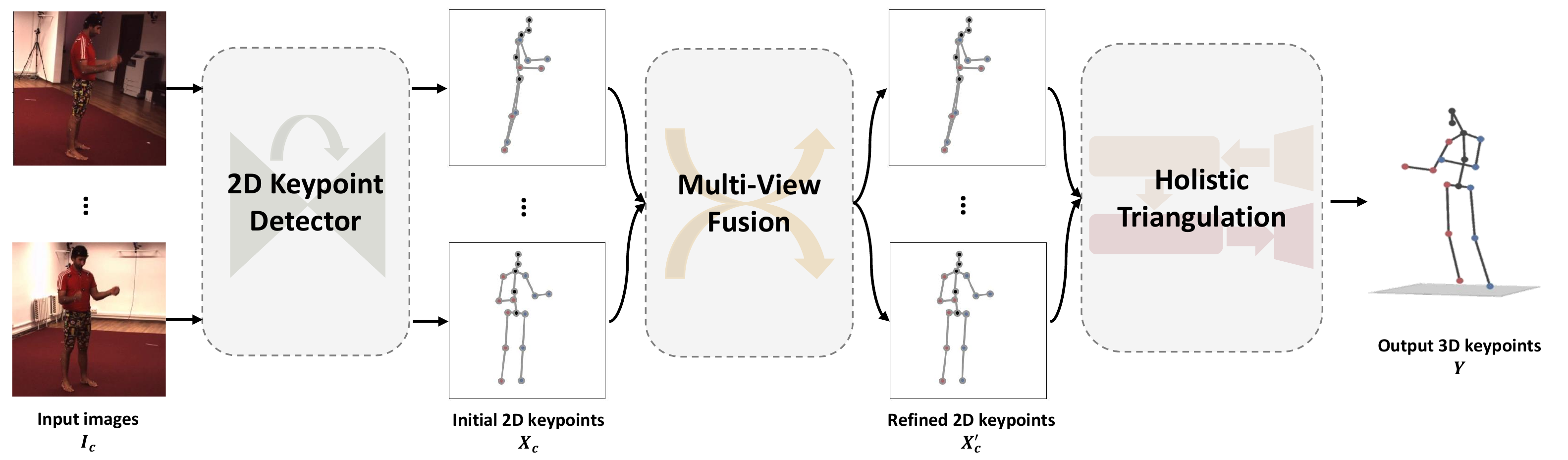}
\vspace{-1em}
\caption{\textbf{The framework of our approach.} 2D keypoint detector achieves 2D poses from multi-view RGB images. MVF refines the 2D results considering the views consistency. And HT generates the 3D pose under the constraints of the anatomy coherence.}
\label{fig:method_framework}
\vspace{-1em}
\end{figure*}

% method
Then to boost the plausibility of 3D poses, Holistic Triangulation (HT) with anatomy constraints is proposed, which enables all 3D keypoints to gain access to pose coherence through 2D-3D phase. Firstly, we modify the formulation of objective function so that all joints can be inferred as an entirety.
% , which is different from the previous methods where each 3D joint is individually estimated. 
% In our method, all 3D joints are concatenated into one vector, and a partitioned diagonal multi-view projection matrix is used to project the whole 3D pose onto 2D image planes. Eventually the final 3D pose, under HT, is found by optimization. 
% However, because of the blockwise linear independence in the objective function, the joint correlation is not well established yet. 
Then, to model the joints linear dependence in the objective function, a PCA reconstruction term is injected. By doing so, joints are coupled in an abstract PCA subspace spanned by the principle components, which contains the global context of whole pose. Human anatomy prior therefore is implicitly introduced. Furthermore, to make the prior more explicit, PCA feature is extended  from keypoint position to skeletal structure feature by applying kinematic chain space (KCS)~\citep{wandt2018kinematic}. Benefiting from the linear property of PCA, HT is still closed-optimized and differentiable.

Consequently, we integrate the 2D detector, MVF and HT into one end-to-end framework and introduce reprojected loss, bone length loss and joint angle loss to promote the view consistency and anatomy coherence during training procedure. In addition, a plausible-pose evaluation metric is proposed to fill in the gap of pose plausibility criterion. 
% A reasonable pose is delineated as containing suitable bone length and limited joint angle.
% The trained occupancy matrix is chosen to model the distribution of joint angle for its facility to couple the azimuth angle and polar angle of one joint. And the training set is composed of Human3.6M~\citep{ionescu2013human3} and MPII-INF-3DHP~\citep{mehta2017monocular}, including about 2.8M frames and covering various motions from normal to challenging.

% performance
Without bells and whistles, MVF-HT method exhibits competitive performance with state-of-the-art techniques, surpassing them in both precision, plausibility and generalization. Moreover, the anatomy prior extracted by PCA is explored through visualization.
% conclusion
The main contributions are summarized below:
\begin{itemize}
    \item We propose a novel MVF module to enhance the view consistency in 2D keypoint estimation. MVF refines 2D keypoint $p$ through perceiving the possible position the same keypoints in other views may localize in the view of $p$.
    \item To our best knowledge, this is the first work to reconstruct the whole 3D pose at once under the triangulation framework. Besides, we inject the anatomy prior extracted by PCA to restrict the pose coherence. In this way, the plausibility of pose is improved.
    \item Our framework can be trained end-to-end but without any learning cost in 2D-3D phase because of the closed-form solution of HT. And a plausible-pose evaluation metric is proposed to fill in the gap of pose plausibility criterion.
\end{itemize}
% In we point out a novel direction in aggregating geometry and anatomy constraints in 2D-3D phase, which is worthwhile to devote much efforts.

\section{Related Work}\label{}
\textbf{Multi-View 3D HPE.} 
The current multi-view 3D HPE methods can be divided into two categories according to the aforementioned two steps. The first category focuses on enhancing the 2D pose estimator. In ~\citep{qiu2019cross,he2020epipolar,remelli2020lightweight}, 2D detectors are enabled to perceive 3D information in the process of 2D detection, where~\citep{qiu2019cross} and~\citep{he2020epipolar} use the epipolar constraints to fuse features of corresponding views while~\citep{remelli2020lightweight} directly generate a canonical representation using convolution network. Our MVF is similar to ~\citep{qiu2019cross} and~\citep{he2020epipolar}, but uses the results of 2D detector to sample joint feature and generate the corresponding pseudo heatmap to provide the assistant for the reference view.

The other category focuses on the second procedure which lifts 2D keypoints to 3D poses. The approach can be summarized as learning-based and optimization-based. In~\citep{iskakov2019learnable,dong2019fast,remelli2020lightweight,kocabas2019self}, based on the camera projection geometry and multi-view 2D points, triangulation~\citep{2003Multiple} is used  to obtain 3D results by SVD or Least-Square method. In~\citep{remelli2020lightweight}, a lightweight DLT method is proposed and exceeds the SVD in time cost. In~\citep{kadkhodamohammadi2021generalizable}, triangulation is replaced with a convolutional network to learn the lifting process. In~\citep{burenius20133d,pavlakos2017harvesting,qiu2019cross}, the human skeleton is modeled as 3D-PSM to establish the potential function combining the 2D observation and skeletal bone length constraints. 3D convolution is applied in~\citep{iskakov2019learnable,tu2020voxelpose} to make the inference directly from a volume. Where the volume is aggregated by multi-view 2D features. PSM and learning-based methods have disadvantages in high computing and time consumption. Conventional triangulation methods only utilize the observation information and geometric constraints but ignore the skeletal prior. Our work not only inherits the cost advantage of triangulation but also injects anatomy prior to maintain the pose coherence. 

\begin{figure*}[!t]
\centering
\includegraphics[width=0.85\textwidth]{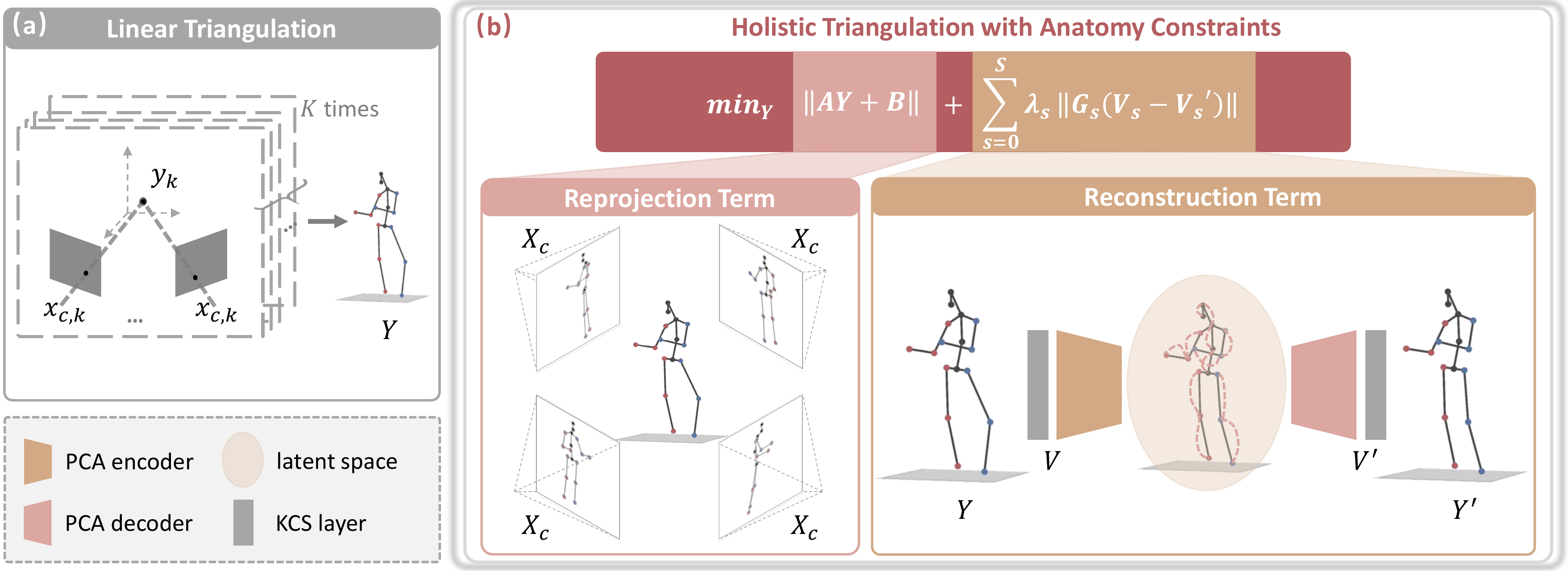}
\vspace{-0.5em}
\caption{\textbf{The schematic diagram of (a) Linear Triangulation (LT) and (b) our Holistic Triangulation (HT) with anatomy constraints.} There are two major differences between two methods: (1) LT reconstructs 3D keypoints separately and concatenates all keypoints to a pose, while HT reconstructs an entire 3D pose at once. (2) LT only consider about the geometric constraints, however, HT includes anatomy constraints extracted by PCA encoder-decoder.}
\label{fig:method_HT}
\vspace{-1em}
\end{figure*}

\textbf{Anatomy Prior Extraction.}
% There have been many attempts in extracting human anthropometric prior to improve joint prediction. 
The prior extraction can be classified as model-based and learning-based.~\citep{zhou2016sparseness} uses the basis pose as the dictionary to represent the pose prior.~\citep{bogo2016keep} employs the SMPL model to limit the result. Although the model brings strong constraints, the iterative optimization used to solve the dictionary weights or SMPL parameters is time-consuming. 
GCN~\citep{cai2019exploiting,liu2020learning,zhao2019semantic} and Attention~\citep{GUO2021103278} are used to capture the relationship between two joints. 
% In~\citep{zeng2021learning}, a hierarchical fusion layer is proposed to learn the association information from distant nodes. 
In~\citep{yang20223d,chen2019weakly,habibie2019wild},  the encoder-decoder is applied to create the latent space which is used to mine the inter-dependencies between joints.
GAN is another kind of model to capture the distribution of poses~\citep{tian2021adversarial,wandt2019repnet,chen2019unsupervised}.
Learning-based methods leverage the power of deep network to capture more generic constraints, but at the cost of more computing resources and network complexity.  In~\citep{malleson2020real}, PCA is employed as a dimensionality-reduction method to acquire pose prior. The linear and network independent properties of PCA attract us. Hence, we use PCA with skeletal structure features as input to learn the relationships from near and distant joints.

\section{Methodology}\label{}
The overview of the proposed method is depicted in Fig.~\ref{fig:method_framework}. There are three major modules: (1) 2D Keypoint Detector, to detect multi-view 2D joint locations respectively, where an off-the-shelf ResNet-152 backbone~\citep{xiao2018simple} is directly applied. (2) Multi-View Fusion (MVF), to refine 2D poses considering the view consistency. (3) Holistic Triangulation (HT), to reconstruct the final 3D pose by closed-form optimization.

The input to the whole framework is a set of multi-view RGB images $I_{c}$, whose index is the number of the synchronized cameras and $c\in\{1,2,\cdots,C\}$. And the output is 3D pose $Y=[\mathbf{y}_{1}^{T},\mathbf{y}_{2}^{T},\cdots,\mathbf{y}_{K}^{T}]^{T}\in\mathbb{R}^{(3K,1)}$, where $\mathbf{y}_{k}=[x_{k}, y_{k},z_{k}]^{T}$ and $K=17$. Each image will be fed into the 2D detector to generate the initial 2D pose $X_{c}=[\mathbf{x}_{(c,1)}^{T},\mathbf{x}_{(c,2)}^{T},\cdots,\mathbf{x}_{(c,K)}^{T}]^{T}\in\mathbb{R}^{(2K,1)}$, where $\mathbf{x}_{c,k}=[x_{(c,k)}, y_{(c,k)}]^{T}$ is the location of the $k^{th}$ joint in view $c$. Then, the MVF module obtains the refined 2D poses $X^{'}_{c}$ from the initial ones by fusing all heatmaps corresponding to different views. After that, HT reconstructs the 3D pose $Y$ from the refined 2D poses through optimization. Finally, a loss function, takes multi-view consistency and whole pose coherence into account, supervises the network when end-to-end training.
%It consists of two parts: reprojection term and reconstruction term. In the first term, the whole 3D pose is reprojected onto image planes, making the pose an entirety and considering geometric constraints. In the reconstruction term, the estimated pose is encouraged to be close to the PCA reconstructed pose $Y^{'}$ which is regarded as plausible as perceiving anatomy coherence.  Furthermore, to enhance the interpretability of the anatomy coherence, KCS layer is capitalized to the PCA 3D pose input to draw joint-related features $V$.

In this section, we first introduce HT, since the goal of our task is 3D pose. Then, the MVF is introduced as an assistance to refine the 2D results. Finally, the overall loss function of the end-to-end framework will be present.

\subsection{Holistic Triangulation with Anatomy Constraints}
% LT a classic optimization method to reconstruct absolute 3D position of one point from its multi-view 2D locations. The target of LT is to minimize the sum of errors between estimated 2D keypoints $\mathbf{x}_k$ and reprojected ones in all views. To infer the 3D pose, as depicted in Fig.~\ref{fig:method_HT}(a), LT is utilized to generate 3D position $\mathbf{y}_k$ of each keypoint separately for $K$ times, and the keypoints are concatenated into one pose under a fixed order. LT is elegant because of the closed-form solution but lack of the joint-by-joint relation modeling. 
LT is classic and elegant because of the closed-form solution, but is lack of joint-by-joint relation modeling. As depicted in Fig.~\ref{fig:method_HT}(a), LT infers 3D position $\mathbf{y}_k$ of each keypoint separately for $K$ times, and the keypoints are then concatenated to generate the 3D pose.
To fix this issue, we propose Holistic Triangulation (HT), shown in Fig.~\ref{fig:method_HT}(b), to reason the whole pose at once through reprojection and reconstruction term:
\begin{equation}
    \min \limits_{Y} \quad  \Vert A Y + B\Vert + \mathbb{H}
\label{eq:method_holistic_triangulation}
\end{equation}
\vspace{-1.5em}
\begin{small}
\begin{equation}
 A=
\begin{bmatrix}
\mathbf{w}_1\circ A_1 &O &\cdots &O\\
O &\mathbf{w}_2\circ A_2 &\cdots &O\\
\vdots &\vdots &\vdots &\vdots\\
O &O &\cdots &\mathbf{w}_k\circ A_k
\end{bmatrix}, 
\quad
% Y=
% \begin{bmatrix}
% \mathbf{y}_1\\
% \mathbf{y}_2\\
% \vdots\\
% \mathbf{y}_K\\
% \end{bmatrix},
% \quad
B=
\begin{bmatrix}
\mathbf{w}_1 \circ \mathbf{b}_1\\
\mathbf{w}_2 \circ \mathbf{b}_2\\
\vdots\\
\mathbf{w}_k \circ \mathbf{b}_k\\
\end{bmatrix}
\nonumber
\end{equation}
\end{small}
where $\circ$ is the Hadamard product. $A_k\in\mathbb{R}^{(2C,3)}$ is the first three columns of $\widetilde{A}_k$ and $\mathbf{b}_k\in\mathbb{R}^{(2C,1)}$ is the last column. $\widetilde{A}_k$ is same as LT (refer to supplementary). We draw the idea from Algebraic Triangulation (AT)~\citep{iskakov2019learnable} to add the learnable confidence $\mathbf{w}_k=[\omega_{1,k}, \omega_{1,k}, \cdots, \omega_{C,k}, \omega_{C,k}]^{T}$ to mitigate the impact of 2D positions with low confidence.

However, owing to the blockwise linear independence in the reprojection term of Eq.~\ref{eq:method_holistic_triangulation}, resulting vector of each block in $Y$ has no difference from AT. To solve this problem, we introduce a reconstruction term $\mathbb{H}$. 
% We firstly leverage the vanilla PCA encoder-decoder to reconstruct a pose $Y^{'}$ from $Y$. Through approaching $Y$ to $Y^{'}$, the linear dependencies of different keypoints are added. Furthermore, skeletal structure features are extracted by KCS~\citep{wandt2018kinematic}, add artificial auxiliary to compensate for the abstraction of PCA.
\begin{figure*}[t]
\begin{minipage}[t]{0.25\linewidth}
    \centering
    \includegraphics[width=0.47\textwidth]{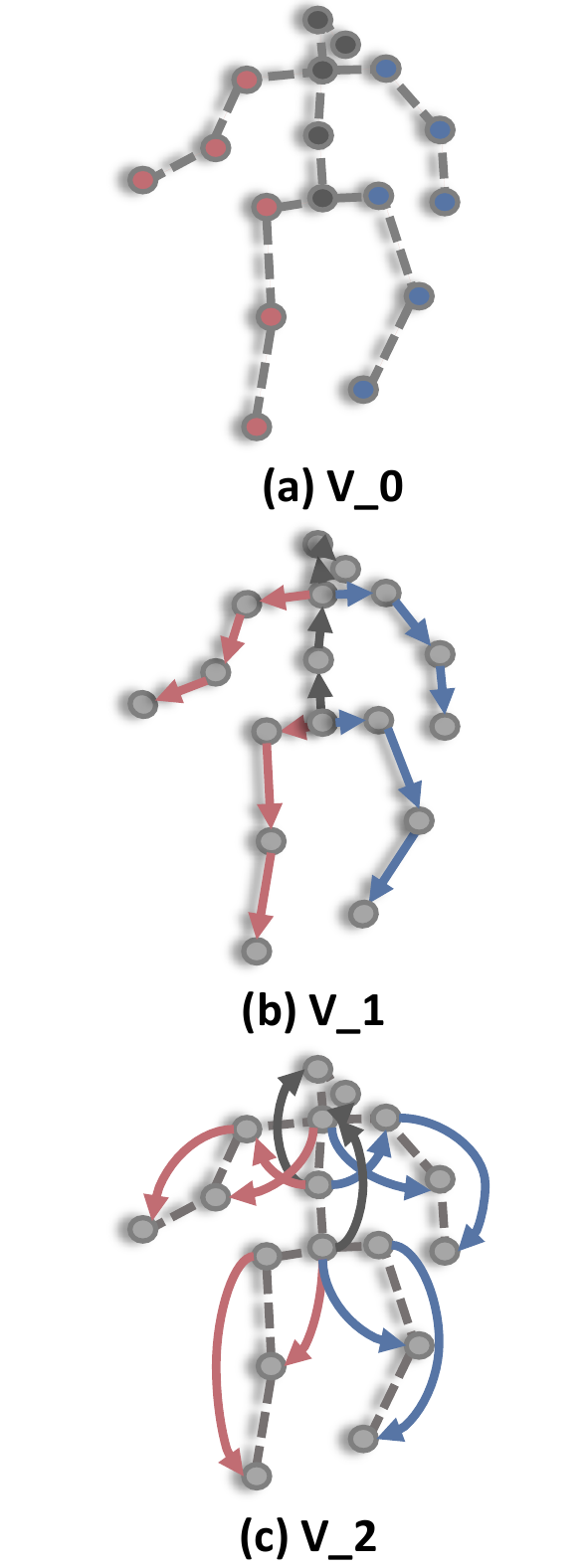}
    % \vspace{1em}
    \caption{Skeletal structure features.}
    \label{fig:method_skeletal_structure}
\end{minipage}
\hspace{1ex}
\begin{minipage}[t]{0.75\linewidth}
    \centering
    \includegraphics[width=\textwidth]{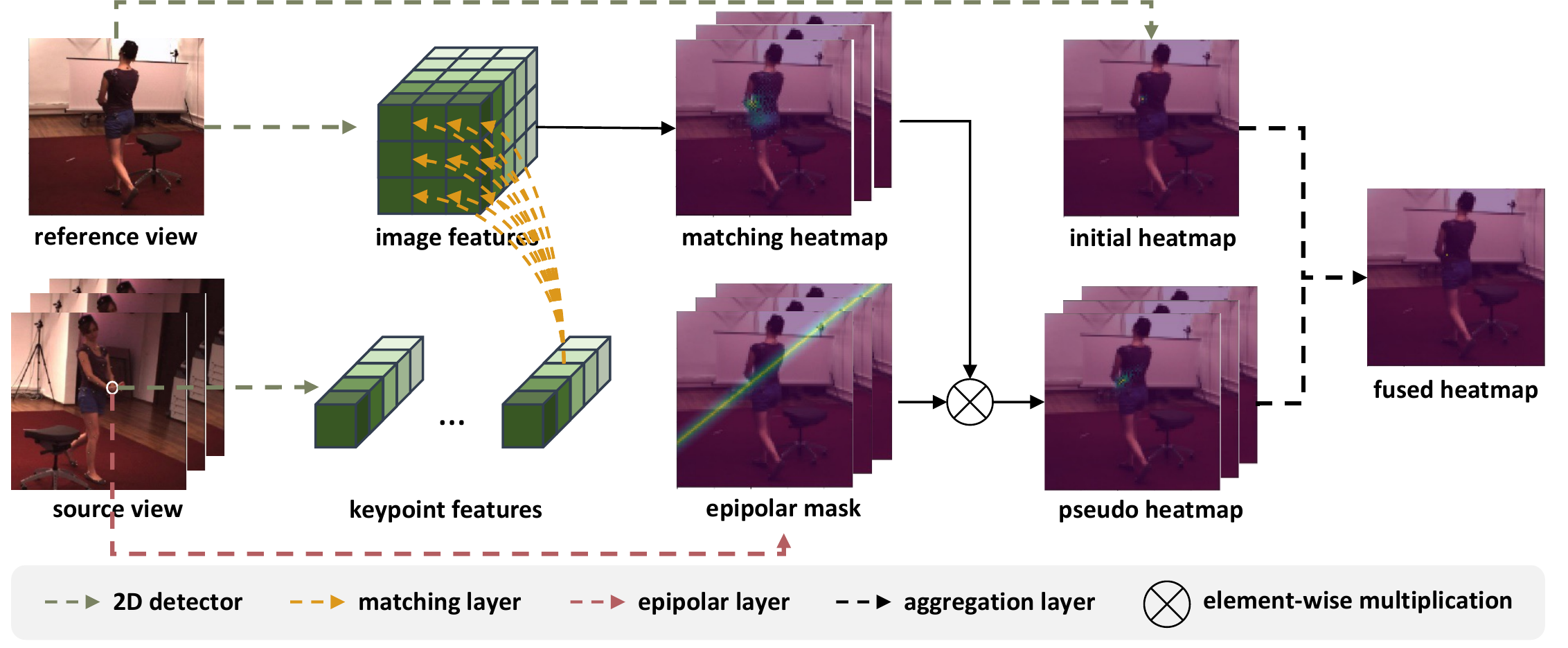}
    \vspace{-2em}
    \caption{\textbf{The pipeline of MVF module.} The reference view image provides the image features and initial heatmap while source views provide keypoint features. Matching heatmap is generated through a matching layer by comparing keypoint features with image features. And it is multiplied by an epipolar mask to avoid mismatching. Finally we aggregate initial heatmap of reference view and the pseudo heatmaps from source views to a fused heatmap.}
    \label{fig:method_MVF}
\end{minipage}
% \vspace{1ex}   %%两个minipage之间相隔3个字符的距离
\vspace{-1em}
\end{figure*}

\noindent\textbf{Vanilla Reconstruction Term.}
The reconstruction term aims to enhance the blockwise linear dependence in $A$ and inject anatomy coherence to $Y$. PCA~\citep{hotelling1933analysis}, a simple but effective module is chosen to model the anatomy prior for two major reasons: (1) The PCA low-dimension latent space is capable to extract the correlations between different keypoints and factor out the generic pose global context. The generality of the pose context is guaranteed by the fact that training data contains various motions. And then we approach the estimated pose $Y$ close to the PCA recovered pose $Y^{'}$ from the latent space, to inject the pose prior. (2) The linear property of PCA will not change the closed-form solution superiority of HT, which will not hinder end-to-end training.

Note that the training set of PCA is root-relative, $Y_{re}=Y-Y_{root}\in\mathbb{R}^{(3K,1)}$. 
% otherwise the scope will be too large to find generic information. 
And $Y_{root}$ is the pelvis position which is estimated by LT. By adding a reconstruction term, the objective function is expressed as:
\vspace{-0.5em}
\begin{equation}
    \min\limits_{Y} \quad  \Vert A Y + B\Vert + \lambda \Vert Y_{re} - Y_{re}^{'} \Vert
\label{eq:method_holistic_triangulation_with_prior}
\end{equation}
where $Y_{re}^{'} = M^{T}M(Y_{re}-Y_{mean})+Y_{mean}$ is the recovered pose; $M\in\mathbb{R}^{(D,3K)}$ is the feature extracting matrix of PCA encoder; $Y_{mean}\in\mathbb{R}^{(3K,1)}$ is the mean pose of PCA training set; and $\lambda$ is a learnable weight of reconstruction term.
Because of the convexity of Eq.\ref{eq:method_holistic_triangulation_with_prior}, the 3D pose can be closed-form solved using Least-Square method (see supplementary for proof):
% add proof in supplementary material
\vspace{-0.5em}
\begin{equation}
    (A^{T}A+\lambda N^{T}N)Y = \lambda N^{T}N(Y_{root}+Y_{mean})-A^{T}B 
    % \nonumber 
    % \\ 
    % Y =(A^{T}A+\lambda N^{T}N)^{-1} (\lambda N^{T}N(Y_{root}+Y_{mean})-A^{T}B)
\label{eq:method_holistic_triangulation_with_prior_closure_answer}
\end{equation}
where $N=I-M^{T}M\in\mathbb{R}^{(3K,3K)}$.
% As illustrated in Fig.~\ref{fig:method_pca_reconstruction_term}, the reprojection term of HT encourages the estimated pose to approach the reprojected pose satisfied with geometric constraints; and the reconstruction term eliminates implausible results unsatisfied with the anatomy coherence and searches for real pose, acting as the anatomy prior to restrict solution domain. 
% In our experiments, different choices of low dimension $D$ reserved by PCA are evaluated empirically, and the ability to extract the prior of reconstruction term is demonstrated.
%\input{method/fig_pca_and_skeletal_structure}

\setlength{\tabcolsep}{2pt}
\begin{table*}[!t]
\definecolor{mygray}{gray}{.9}
\begin{center}
\caption{\textbf{Comparison with state-of-the-art methods on Human3.6M in terms of 2D pose estimation accuracy metric JDR (\%).} ``dot'' means using inner dot matching strategy and ``fcl'' represents fully connected layer.}
\vspace{0.5em}
\label{tb:exp_compare_with_sota_2d_h36}
\resizebox{0.7\textwidth}{!}{
\begin{tabular}{lcccccccccccc}
\toprule
Method    & shlder        & elb           & wri           & hip           & knee          & ankle         & root          & belly         & neck          & nose          & head          & Avg.          \\ \midrule
CrossView~\citep{qiu2019cross} & 95.6          & 95.0          & 93.7          & 96.6          & 95.5          & 92.8          & 96.7          & 96.4          & 96.5          & 96.4          & 96.2          & 95.9          \\ 
Epipolar~\citep{he2020epipolar}  & 97.7          & 97.3          & 94.9          & \textbf{99.8} & 98.3          & 97.6          & \textbf{99.9} & \textbf{99.9} & \textbf{99.8} & 99.7          & \textbf{99.5} & 98.3          \\ \rowcolor{mygray}
\textbf{ours-dot}  & 96.4          & 96.8          & \textbf{99.8} & 97.2          & 98.3          & 99.5          & 97.3          & 99.7          & \textbf{99.8} & 99.6          & 93.7          & 98.0          \\ \rowcolor{mygray}
\textbf{ours-fcl}  & \textbf{97.8} & \textbf{97.5} & \textbf{99.8} & 97.7          & \textbf{98.7} & \textbf{99.6} & 97.8          & 99.7          & \textbf{99.8} & \textbf{99.8} & 95.4          & \textbf{98.5} \\ \bottomrule
\end{tabular}
}
\end{center}
\vspace{-2.5em}
\end{table*}
\setlength{\tabcolsep}{1.4pt}
\setlength{\tabcolsep}{2pt}
\definecolor{mygray}{gray}{.9}
\begin{table*}[t]
\begin{center}
\caption{\textbf{Comparison with state-of-the-art methods on Human3.6M in terms of MPJPE,} where the input of 2D-3D step is 2D locations. T. is short for triangulation. }
\label{tb:exp_compare_with_sota_h36}
% \vspace{-1em}
\resizebox{0.95\textwidth}{!}{
\begin{tabular}{lcccccccccccccccc}
\toprule
MPJPE (mm)                    & Dir. & Disc. & Eat   & Greet & Phone  & Photo & Pose   & Purch. & Sit  & SitD. & Smoke & Wait  & WalkD. & Walk  & WalkT. & Avg.\\ \midrule
Canonical~\citep{remelli2020lightweight}  &27.3 &32.1 &25.0 &26.5 &29.3 &35.4 &28.8 &31.6 &36.4 &31.7 &31.2 &29.9 &26.9 &33.7 &30.4 &30.2  \\
CrossView-T.~\citep{qiu2019cross}           &25.2 &27.9 &24.3 &25.5 &26.2 &23.7 &25.7 &29.7 &40.5 &\textbf{28.6} &32.8 &26.8 &26.0 &28.6 &25.0 &27.9  \\
Epipolar-T.~\citep{he2020epipolar} &29.0 &30.6 &27.4 &26.4 &31.0 &31.8 &26.4 &28.7 &34.2 &42.6 &32.4 & 29.3 &27.0 &29.3 &25.9 &30.4\\

Algebraic-T.~\citep{iskakov2019learnable}        & 20.4 & 22.6  & 20.5  & 19.7  & 22.1   & 20.6  & 19.5   & 23.0  & 25.8 & 33.0    & 23.0    & 21.6  & 20.7   & 23.7  & 21.3   & 22.6   \\ 
\rowcolor{mygray} \textbf{ours-MVF}                 & 20.1 & 21.5  & 20.0   & 18.7 &21.3  & 20.3  & 18.4 & 21.9  & 24.3 & 30.6  & 22.1  & 20.4 &19.6   & 23.4 & 20.2   & 21.6\\
\rowcolor{mygray} \textbf{ours-HT}                 & \textbf{19.4} & 21.5  & 20.0   & 18.7 &21.6  & 20.9  & 18.2 & 21.5  & 24.8 & 31.7  & 21.7  & 20.2 &\textbf{18.9}   & 23.2 & 19.6   & 21.6\\
\rowcolor{mygray} \textbf{ours}                 & 19.5 & \textbf{20.9}  & \textbf{19.5}   & \textbf{18.3}  & \textbf{21.1}  & \textbf{20.0}  & \textbf{17.9} & \textbf{21.3}     & \textbf{23.9} & 30.1  & \textbf{21.6}  & \textbf{19.9} & \textbf{18.9}   & \textbf{22.8} & \textbf{19.5}   & \textbf{21.1}\\ \bottomrule
\end{tabular}}
\end{center}
\vspace{-2em}
\end{table*}
\setlength{\tabcolsep}{1.4pt}

\noindent\textbf{Skeletal Structure Feature Extraction Module.\label{sec:method_skeletal_structure_feature_extraction_module}}
One inadequacy of the basic reconstruction term above is that the prior extracted is implicit. To address this, we transform the data from keypoint space to skeletal structure space, enhancing associations between joints and introducing explicit features.

To generate skeletal structure feature $V$, KCS, a matrix multiplication algorithm to create vector between two selected points, is used to transform joints to joint-connected vectors by a mapping matrix $C \in \mathbb{R}^{(3J,3K)}$. And $G \in \mathbb{R}^{(3K,3J)}$ is applied to transform back.

The feature of connected joint with $s$ hops is named as $V_{s}$. As shown in Fig.~\ref{fig:method_skeletal_structure}, both keypoints and bone vectors can be compatible with $V\_0$ and $V\_1$. Because longer distances will result in fewer extracted vectors with less information, only $hop=0,1,2$ is defined.
% First, considering the topology of human skeleton, we modify PCA input data from the keypoint coordinates $Y_{re}$ to bone vectors $V\in \mathbb{R}^{(3J,1)}$ ($J$ is the number of bones). 
% Obtaining bone vectors, the reconstruction term in Eq.~\ref{eq:method_holistic_triangulation_with_prior} is adapt to $\lambda \Vert V - V^{'} \Vert$.
% $V^{'}$ is the recovered bone vector by PCA.
% In order to keep the triangulation projection term and PCA reconstruction term in the same dimension, we utilize Eq.\ref{eq:method_kcs_verse} to remap the reconstruction error. Consequently, the objective function of HT can be written as:
% \input{method/eq_holistic_triangulation_with_kcs}
% Then inspired by the observation in~\citep{zeng2021learning} that the information provided by distant joints is also useful, we extend the feature to joint-connected space extracted from two k-hop neighbor joints.  
By fusing different $V\_s$ features, 
% the reconstructed term is $\sum_{s=0}^{S}\lambda_{s} \Vert V_{s} - V_{s}^{'} \Vert$, and 
the objective function can be adapted to:
\vspace{-0.5em}
\begin{equation}
    \min\limits_{Y} \quad  \Vert A Y + B\Vert + \sum_{s=0}^{S}\lambda_{s} \Vert G_{s} (V_{s} - V_{s}^{'})\Vert
\label{eq:method_holistic_triangulation_with_multi_kcs}
\end{equation}
where $G_{s}$ remaps the reconstructed error from feature space back to keypoint space in order to keep two terms in the same dimension.
Replacing $V$ with $CY$ and the solution is:
\vspace{-0.5em}
\begin{small}
\begin{equation}
    (A^{T}A+ \sum_{s=0}^{S} \lambda_{s} H_{s}^{T}H_{s})Y =\sum_{s=0}^{S} \lambda_{s} H_{s}^{T}H_{s}(Y_{root}+Y_{mean})-A^{T}B
\label{eq:method_holistic_triangulation_with_multi_kcs_closure_answer}
\end{equation}
\end{small}
where $H_{s}=G_{s}N_{s}C_{s}\in\mathbb{R}^{(3K,3K)}$. 
% We assess aggregation strategies in the experiment. 

\subsection{Multi-View Fusion 2D Keypoint Refinement}
The initial 2D keypoints achieved by 2D backbone detector are independent from each view. To enhance the cross view correlations, we introduce the MVF module. The pipeline of MVF is illustrated in Fig.~\ref{fig:method_MVF}. To make the keypoints in the reference view consistent with other views, the pseudo heatmaps corresponding to the same keypoints in other views are generated. Concretely, the pseudo heatmap is the product of matching heatmap and epipolar mask, and
represents the probability source view keypoint localizing in the reference view.
% And the matching heatmap is the probability map which measures the matching degree of the keypoint features in source view and the whole image features in reference view. 
After that, the initial heatmap are fused with pseudo heatmaps through an aggregation layer which is a $1*1$ convolution kernel to product the refined fused heatmap. 

% In the rest of this section, we will introduce the detailed process of matching layer and epipolar layer.
% \begin{figure*}[t]
% \centering
% \includegraphics[width=0.85\textwidth]{method/MVF_v2.0.pdf}
% \vspace{-1em}
% \caption{\textbf{The pipeline of MVF module.} The reference view image provides the image features and initial heatmap while source views provide keypoint features. Matching heatmap is generated through a matching layer by comparing keypoint features with image features. And it is multiplied by an epipolar mask to avoid mismatching. Finally we aggregate initial heatmap of reference view and the pseudo heatmaps to a fused heatmap.}
% \label{fig:method_MVF}
% \vspace{-1em}
% \end{figure*}
\noindent\textbf{Matching Layer.}
% Ideally, in the reference view, for one 2D keypoint $p$, if we know the corresponding locations $p_{pseudo}$ projected from same keypoints $p^{'}$ in source views. All we need to do is offsetting the $p$ to $p_{pseudo}$ to enhance the consistency between $p$ and $p^{'}$. However, there is no geometric way to project one point in one view to the other view. 
The idea of cost volume in stereo matching methods~\citep{kendall2017end,xu2020aanet} inspires us. 
% The disparity of two same points in different views are regressed from the cost volume. 
% The cost volume measures the similarity degree between two corresponding points under the fixed disparity. 
The corresponding matching heatmaps $H_{match}$, which indicates the matching degree of the keypoints $p^{'}$ in the source view and all pixels $p(i,j)$ in the reference view, is generated. The pixel gets higher matching score as its features are better matched with $p^{'}$.
% : 
% \input{method/eq_matching_heatmap}
% Following this, a convolution layer is added to smooth the matching score. 
We also explore two types of matching strategy:
\begin{itemize}
    \item inner dot: $\frac{1}{N} ( F(i,j) \cdot F(p^{'}))$
\vspace{-0.5em}
    % \item element-wise deviation: $\frac{1}{N} \sum \limits_{n} (F(i,j)_{n} - F(p^{'})_{n})$
    \item fully connected layer: $\mathbf{w}^{T}\cdot cat(F(i,j),F(p^{'}))$
\end{itemize}
where $F$ represents the features generated by 2D backbone, $F(p^{'})$ is the sampled feature of $p^{'}$ via bilinear interpolation, $cat()$ means concatenation and $\mathbf{w}$ is the learnable parameters.
% To consider more information and improve the tolerance of $p_{pseudo}$ mistake. Instead of offsetting the points, we fuse the heatmaps through an aggregation layer and then utilize the soft-argmax to calculate the refined 2D keypoint location from the fused heatmap.

\noindent\textbf{Epipolar Layer.}
In stereo matching task, only the points in the horizontal direction will be compared because the given image pair are rectified. However, in matching layer, the $p^{'}$ is compared with all pixels in the reference view due to lack of rectification. The matching instability will be caused by the similar feature vectors of wrong pixels. To solve the problem, we generate the epipolar mask by the epipolar field~\citep{ma2021transfusion} to limit the matching pixels locating near the epipolar line of $p^{'}$. The epipolar field indicates the probability pixels $p(i,j)$ in the reference view lies in the epipolar line of $p^{'}$: 
\vspace{-0.5em}
\begin{equation}
    C(p,p^{'}) = (1 - |(\overrightarrow{c^{'}p^{'}}\times\overrightarrow{cc^{'}})\cdot\overrightarrow{cp(i,j)}|)^{\gamma}
\label{eq:method_epipolar_field}
\end{equation}
where $c,c^{'}$ are camera centers and $\gamma$ is the soft factor to control the epipolar field margin. The field gets narrower as $\gamma$ gets bigger, we choose $\gamma=10$ to generate our epipolar mask by empirical results.

% Depending on the epipolar geometry, two camera centres $c$ and $c^{'}$, point $p^{'}$ and its corresponding epipolar line $l$ lie in a common plane $\mathbf{\pi}$. Hence, if point $p(i,j)$ is in the $l$, the plane $\mathbf{\pi}$ must contains the ray $\overrightarrow{cp(i,j)}$. In other words, $\overrightarrow{cp(i,j)}$ is perpendicular to the normal vector $\overrightarrow{c^{'}p^{'}}\times\overrightarrow{cc^{'}}$ of the $\mathbf{\pi}$. The correspondence score is calculated by:
% the inner product of $\overrightarrow{cp(i,j)}$ and the normal vector of $\mathbf{\pi}$:

\subsection{Loss\label{sec:method_loss}}
The overall loss function consists of four parts: (1) Mean Square Error (MSE) between estimated 3D pose and groundtruth, (2) reprojected error: L2 loss of reprojeted 2D pose and estimated 2D pose, (3) bone length loss: L1 loss of estimated bone vector and groundtruth and (4) joint angle loss of 3D poses: 
\vspace{-0.5em}
\begin{small}
\begin{equation}
    L(Y) = L_{MSE}(Y,\hat{Y}) + \beta_{pj}L_{pj}(X^{'}, Y) + \beta_{bl}L_{bl}(Y,\hat{Y})+ \beta_{ja}L_{ja}(Y)
\label{eq:method_loss}
\end{equation}
\end{small}
where $\hat{Y}$ is the groundtruth of the pose; $\beta_{pj}$, $\beta_{bl}$ and $\beta_{ja}$ are set to 0.1, 0.01, 0.01 separately by empirical results.
Based on the common $L_{MSE}$, we subjoin the $L_{pj}$ to enhance the multi-view consistency and $L_{bl}, L_{ja}$ to promote the anatomy coherence.
% \noindent\textbf{Reprojected Loss.}
% The reprojected loss is expressed as the sum of the errors between reprojected keypoints from $Y$ and refined keypoints $X^{'}$:
% \input{method/eq_pj_loss}
% \noindent\textbf{Bone Length Loss.}
% L1 norm is utilized to measure distance of bone length between the solved pose and the annotation:
% \input{method/eq_bl_loss}
% \begin{figure}[t]
% \centering
% \includegraphics[width=0.8\textwidth]{method/jointangledistribution.pdf}
% \caption{Illustration of the distribution of selected joints' angles, the left image is the distribution of azimuth angle and the right one is polar angle. Here, we select nine joints angles to observe which are right-shoulder, right-elbow, left-shoulder, left-elbow, right-hip, right-knee, left-hip, left-knee, and spine from left to right in each image.}
% \label{fig:method_joint_angle_distribution}
% \end{figure}

\noindent\textbf{Joint Angle Loss.}
% Intuitively, most joint angles are restricted in a certain range.
% The angle of the $k^{th}$ joint is defined as the angle between two bones connected to this joint. 
% To calculate joint angle, we convert the global coordinate to  similar as~\citep{akhter2015pose} and get the azimuth $\varphi_{k}$ and polar angle $\theta_{k}$.
The multivariate GMM is used to model joint angle distribution $p(x_{k})$, $ x_k=[\sin{\theta_k}, \sin{\varphi_k}, \cos{\varphi_k}]^{T}$, and $\varphi_{k}, \theta_{k}$ are azimuth and polar angle of joint in a local spherical coordinate system~\citep{akhter2015pose} (details in suppl.).
% which is illustrated in the supplementary. 
% Therefore, we apply the multivariate GMM to model the distributions of joints angles:
% \input{method/eq_multivariate_GMM}
% The multivariate GMM of each selected joint is generated by EM algorithm~\citep{dempster1977maximum}. 
And the joint angles with low probability are penalized:
\vspace{-0.5em}
\begin{equation}
    L_{ja} = \frac{1}{K_{se}}\sum\limits_{k=0}^{K_{se}} sigmoid \left( \left( p(x_k)-\frac{a}{2} \right)*(-\frac{10}{a}) \right)
\label{eq:method_ja_loss}
\end{equation}
where $K_{se}$ is the number of selected joints; $a$ is probability border $p(x_k\pm3\sigma)$, the angle with the probability $(0,a)$ should be penalized. So transformation $-\frac{a}{2}$, coefficient $\frac{10}{a}$ are used to offset $(0,a)$ to $(5,-5)$ to suit the variable domain of sigmoid.

\section{Experiments}\label{}
\subsection{Datasets and Evaluate Metrics}
% We conduct experiments and validate our approach on two popular multi-view benchmark datasets: Human3.6M~\citep{ionescu2013human3} and Total Capture~\citep{trumble2017total}.

\noindent\textbf{Human3.6M Dataset.} 
The Human3.6M~\citep{ionescu2013human3} is one of the most universal 3D HPE dataset with 3.6 million annotations. The videos are acquired from 4 synchronized cameras in laboratory. 
% , and labels are collected by marker-based motion capture systems. 
We use Joint Detection Rate (JDR) to evaluate 2D pose, Mean Per Joint Position Error (MPJPE) to evaluate relative 3D pose and Percentage of Plausible Pose (PPP), elaborated in Sec.~\ref{sec:dataset_ppp_metric}, to assess plausibility.

\noindent\textbf{Total Capture Dataset.}
The Total Capture Dataset~\citep{trumble2017total} is a common dataset recorded by 8 cameras which are distributed over different pitch angles from top to bottom. The dataset contains various actions, including some challenging motions like crawling and yoga. Hence, cross-dataset experiments are executed on it to evaluate the generalization.

\subsection{Plausible-Pose Evaluation Metric\label{sec:dataset_ppp_metric}}
% To identify a plausible pose, we propose plausible-pose protocol. Based on it, the plausible-pose metric is raised to evaluate the plausibility performance statistically.
\noindent\textbf{Plausible-Pose Protocol.}
A plausible pose should meet two requirements: all bones have appropriate length and all joints are flexed in a limited range. A suitable bone length should be as near as possible to the groundtruth, and a reasonable joint angle is located in an occupancy matrix $OC(\theta, \varphi)$ which indicates whether the angle pair $(\theta, \varphi)$ appears in the training set:
\begin{small}
\begin{equation}
P_{bl}(BL) = \left\{
\begin{array}{lll}
    &1, \quad & |\frac{BL}{\hat{BL}}-1|<R  \\ 
    &0, \quad &others
\end{array}\right.
,
P_{ja}{(\theta,\varphi)} = \left\{
\begin{array}{lll}
    &1, \quad &OC(\theta, \varphi)=1 \\
    &0, \quad &others
\end{array}\right.
\label{eq:exp_r_bl_and_ja}
\end{equation}
\end{small}
where $\hat{BL}$ is bone length groundtruth and $R$ is bone length proportion threshold. The morphology technique is used to smooth the occupancy matrix $OC$, so that the continuous feasibility space can be covered even though the $(\theta, \varphi)$ is discrete.
% where $R$ is the proportion threshold, a reasonable bone length should be as near as possible to the groundtruth. We select $R_{bl}@R$ solved results as logical ones.  

% The reasonable joint angle is located in an occupancy matrix $OC(\theta, \varphi)$ whose value equal to 1 only when the angle pair $(\theta, \varphi)$ appears in the training set: 
Ultimately, a reasonable pose is:  
\vspace{-0.5em}
\begin{equation}
    P_p(Y) = \prod\limits_{j=0}^{J}P_{bl}(BL_{j})\prod\limits_{k=0}^{K_{se}}P_{ja}(\theta_{k},\varphi_{k})
\label{eq:exp_r_pose}
\end{equation}
% where $P_p@R$ is a reasonable pose when $P_{bl}@R$ is logical bone length.

\begin{figure}[!t]
    \centering
    \includegraphics[width=0.45\textwidth]{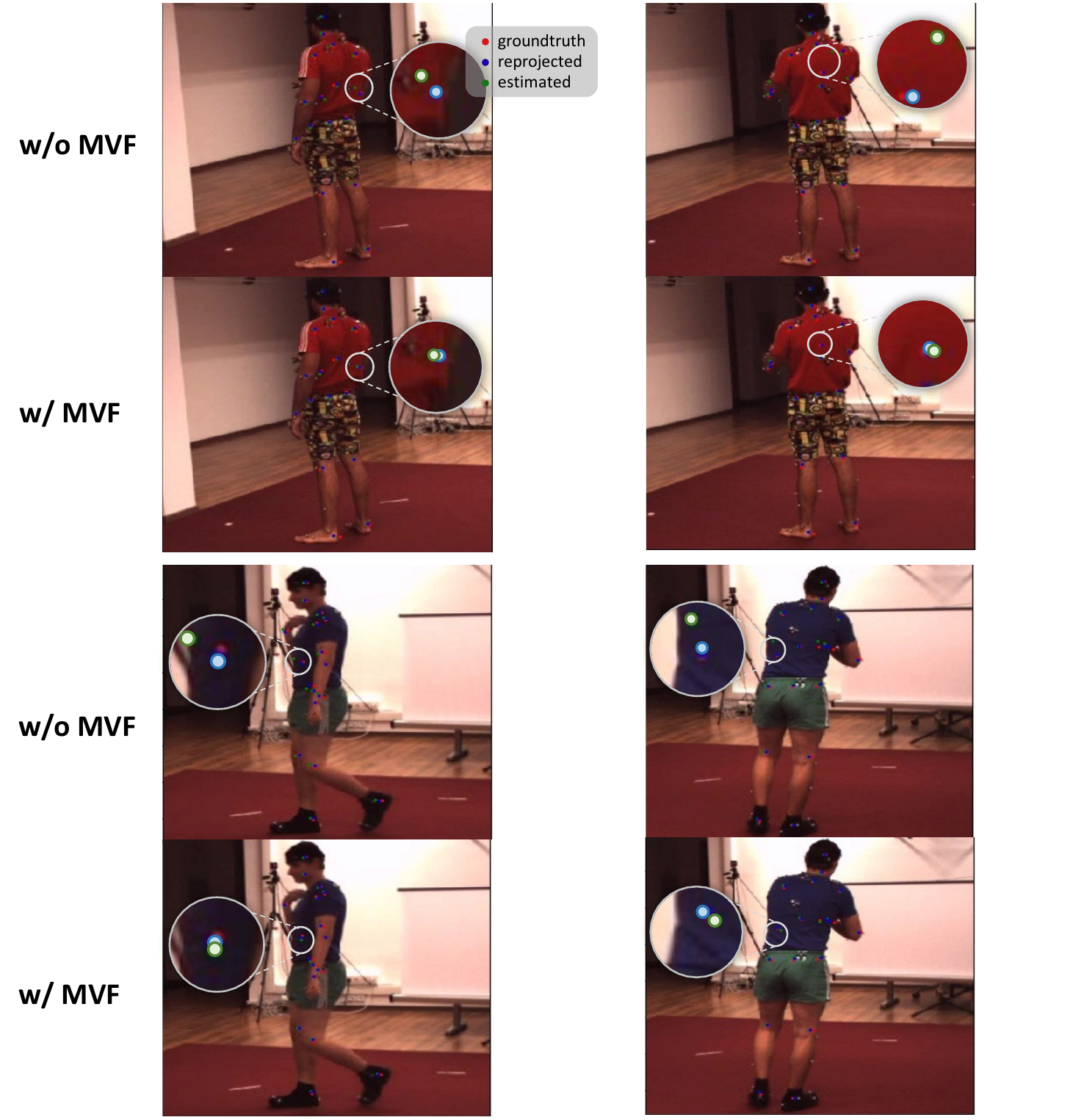}
    \vspace{-1em}
    \caption{\textbf{Visualization of the 2D keypoints with and without MVF refinement.} We use different colors to distinguish different types of 2D results, where red: groundtruth, green: estimates, blue: reprojected results from 3D reconstruction. As reprojected results get closer to estimates, the different view keypoints are more consistent.}
    \label{fig:exp_MVFvisualization}
    \vspace{-1em}
\end{figure}
\noindent\textbf{Plausible-Pose Metric.}
To evaluate the plausibility performance statistically in testing dataset, a new metric PPP is defined as: 
\vspace{-0.5em}
\begin{equation}
PPP = \frac{1}{T} \sum\limits_{t=1}^{T}R_p(Y_t)
\label{eq:exp_ppp}
\end{equation}
where $T$ is the number of testing samples, and the metric is divided according to the bone length threshold into PPP$@R$.

\subsection{Comparison with State-of-The-Art Methods}
We compare quantitative and qualitative performance with the state of the art using all views on Human3.6M, and conduct cross-dataset experiments on Total Capture.

\noindent\textbf{Implementation Details.\label{sec:implementation_detail}}
The low-dimension feature extraction matrix $M$ and mean pose $Y_{mean}$ of PCA are both generated by training set of Human3.6M and MPII-INF-3DHP. In order to avoid the influence of orientations diversity, the orientation normalization is applied in the training data. It should be clarified that hyperparameters and feature extraction strategies are determined by ablation study (supplementary): the feature $V\_0$, $V\_1$, $V\_2$ are fused and the corresponding PCA reserved dimension $D$ are set to $25$, $20$, $15$ respectively; and coefficient of reconstruction term $\lambda$ are learnable with initial value $8000$, $4000$, $4000$. We first train the MVF network with MSE loss of 2D keypoints on the training set of Human3.6M for 2 epochs with a batch size of 12. The learning rate is initially set to $10^{-2}$ and decays every 25000 iterations by a factor of $0.1$. After that, the whole network which combines three modules are trained for 4 epochs on the training set of Human3.6M with $10^{-4}$ learning rate under the supervision of loss funtion described in Sec.~\ref{sec:method_loss}. If not mentioned explicitly, the baseline is AT~\citep{iskakov2019learnable} method.

\begin{figure*}[!t]
    \centering
    \includegraphics[width=0.85\textwidth]{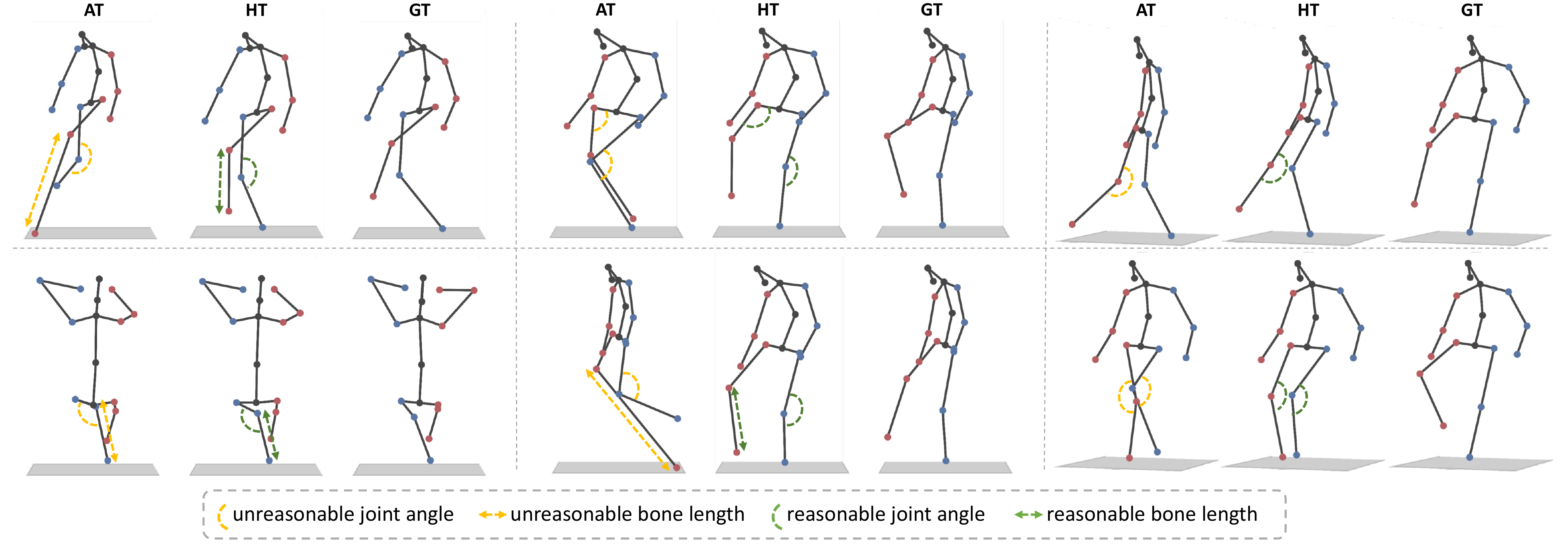}
    \vspace{-1em}
    \caption{Visualization of estimated 3D poses. Different colors are used to distinguish whether the pose is reasonable or not, where yellow represents unreasonable and green is reasonable. HT can amend the unreasonable poses. For example, in the first column of the first row, the pose generated by AT (baseline method, w/o reconstruction term) has a too long right leg (red side) and an unreasonable left knee angle (blue side), which is corrected by HT.}
    \label{fig:exp_qualitative_results}
    \vspace{-1.5em}
\end{figure*}

\noindent\textbf{Quantitative Results on Human3.6M.}
We first evaluate the refined 2D results after MVF refinement module. Following convention, the threshold of JDR is set to the half of the head size.
As shown in Table~\ref{tb:exp_compare_with_sota_2d_h36}, MVF outperforms CrossView by at least $2.1\%$ regardless of the matching strategy. And fcl MVF also surpass Epipolar Transformer. The improvement demonstrates that the initial keypoint location can be leveraged to generate reliable pseudo heatmap with fewer calculations. 

To evaluate the 3D pose estimation, we first compare precision performance with state-of-the-art methods whose input of 2D-3D step is only 2D keypoint locations. In addition to the whole framework, two networks are trained separately: (1) only MVF, uses MSE loss and reprojected loss to supervise, (2) only HT, supervised by MSE loss, bone length loss and joint angle loss. Both proposed modules achieves average MPJPE of $21.6mm$, surpassing AT by $1mm$ (relative $4.4\%$). The improvement demonstrates that view consistency and anatomy coherence are both meaningful for pose estimation. As Table~\ref{tb:exp_compare_with_sota_h36} shows, the method combined with two modules achieves the state-of-the-art results, with $21.1mm$ MPJPE, $6.6\%$ better than AT. And the performance is improved on almost all actions. 

We also compare our approach with other methods whose input of 2D-3D reconstruction is heatmap or intermediate feature which contains 
more infromation than keypoint location. As shown in Table~\ref{tb:exp_compare_with_sota_h36_2}, our method achieves a balance of implementation complexity (calculated by thop\footnote[1]{https://github.com/Lyken17/pytorch-OpCounter}), consuming time and accuracy performance. MVF-HT surpasses AT in both precision and plausibility with $6.6\%$ and $2.4\%$, only at the cost of $30ms$ time consumption. Even though Volumetric Triangulation (VT)~\citep{iskakov2019learnable} surpasses us with $0.3mm$ MPJPE, the MVF-HT almost outperforms it by $300$ billion in the number of operations and $48ms$ in time costing. In 3D reconstruction procedure, VT utilizes 2D features as input and 3D CNN as inference network, which considers more information and is more complicate than ours. 
% And compare AT and VT (two methods in one work but with different input), the advantage of considering more information in 3D reconstruction is obvious. 
For the further work, we will explore the closed-form method that takes 2D features as input. 

\setlength{\tabcolsep}{1.5pt}
\definecolor{mygray}{gray}{.9}
\begin{table}[t]
\begin{center}
    \caption{\textbf{Comparison of MPJPE, inference time, computation complexity and PPP@0.2 on Human3.6M.} MACs and param are shorthand of the number of multiply-add operations and parameters.}
\label{tb:exp_compare_with_sota_h36_2}
\vspace{-0.5em}
\resizebox{0.5\textwidth}{!}{
\begin{tabular}{lccccccccc}
\toprule
\multirow{2}{*}{Method} &
\multicolumn{3}{c}{Input} & \multicolumn{2}{c}{Complexity} &
\multirow{2}{*}{\begin{tabular}[c]{@{}c@{}}MPJPE\\ (mm)\end{tabular}}  & \multirow{2}{*}{\begin{tabular}[c]{@{}c@{}}Time\\ (ms)\end{tabular}}  & \multirow{2}{*}{\begin{tabular}[c]{@{}c@{}}PPP@0.2\\ ($\%$)\end{tabular}}\\
% \cline{2-3} 
&feature &heatmap &keypoint
                  & param           & MACs          &                        \\ \midrule
CrossView-RPSM~\citep{qiu2019cross} &  &\checkmark & & 570M & 424B &26.2 &$1.88\times10^{3}$&-\\
% CrossView-T.~\citep{qiu2019cross} &  & &\checkmark & 570M & 424B &27.9 &110\\
Epipolar-RPSM~\citep{he2020epipolar} &  &\checkmark & & 78M  & 410B &26.9 &$1.88\times10^{3}$
&-\\
% Epipolar-T.~\citep{qiu2019cross} &  & &\checkmark & 78M  & 410B &30.4 &97\\
Algebraic-T.~\citep{iskakov2019learnable}    &  & &\checkmark & 79M   & 418B & 22.6   &\textbf{75}                 &79.36\\
Volumetric-T.\citep{iskakov2019learnable}  &\checkmark  & &  & 80M   & 717B & \textbf{20.8}  &152          &-\\
\rowcolor{mygray} \textbf{ours}  &  & &\checkmark & 79M  &418B  & 21.1  &104                  &\textbf{81.24}\\\bottomrule
\end{tabular}
}
\end{center}
\vspace{-2.5em}
\end{table}

% \vspace{-1em}
% \footnotetext[1]{https://github.com/Lyken17/pytorch-OpCounter}
\setlength{\tabcolsep}{8pt}
\begin{table}[!t]
\begin{center}
    \caption{\textbf{Comparison with the state of the art on Total Capture in terms of MPJPE.} Methods with * are trained on the Total Capture.}
\label{tb:exp_compare_with_sota_tc}
\vspace{-0.5em}
\resizebox{0.5\textwidth}{!}{
\begin{tabular}{lccccccc}
\toprule
\multirow{2}{*}{\begin{tabular}[c]{@{}l@{}}MPJPE\\ (mm)\end{tabular}} & \multicolumn{3}{c}{Subject1,2,3} & \multicolumn{3}{c}{Subject4,5} & \multirow{2}{*}{Avg.} \\ %\cline{2-7}
                                                                      & W2        & FS3        & A3        & W2       & FS3       & A3        &                      \\ \midrule
% Tri-CPM*\citep{wei2016convolutional}                                                               & 79        & 112        & 106       & 79       & 149       & 73        & 99                   \\
% PVH*\citep{trumble2017total}                                                                   & 48        & 122        & 94        & 84       & 168       & 154       & 107                  \\
IMUPVH*\citep{trumble2017total}                                                               & 30        & 91         & 49        & 36       & 112       & \textbf{10}        & 70                   \\
AutoEnc*\citep{trumble2018deep}                                                              & \textbf{13}      & 49         & 24        & \textbf{22}      & 71        & 40        & 35                   \\
CrossView*\citep{qiu2019cross}                                                            & 19        & 28         & 21        & 32       & 54       & 33        & 29                  \\ 
GeoFuse*\citep{zhang2020fusing}                                                            & 14        & 26         & 18       & 24       & 49        & 28       & 25                   \\ 
\rowcolor{mygray} baseline                                                                  & 64        & 60         & 53        & 72       & 78        & 62        & 63   \\
\rowcolor{mygray} ours                                                                  & 45        & 49         & 45       & 52       & 64       & 57       & 51   \\
\rowcolor{mygray} ours*                                                                  & \textbf{13}        & \textbf{24}         & \textbf{17}        & 23       & \textbf{41}        & 29        & \textbf{23} 
\\ \bottomrule
\end{tabular}}
\end{center}
\vspace{-1.8em}
\end{table}
\setlength{\tabcolsep}{1.4pt}
\setlength{\tabcolsep}{7pt}
\begin{table*}[t]
\begin{minipage}{0.3\linewidth}
    \begin{center}
\caption{\textbf{Effect of matching strategy in MVF.} First row is baseline.}
\label{tb:exp_matching_strategy}
% \vspace{-0.5em}
\resizebox{\textwidth}{!}{
\begin{tabular}{cc|cc|c}
\Xhline{0.09em}
\multicolumn{2}{c|}{Matching Strategy} & \multicolumn{2}{c|}{View} & \begin{tabular}[c]{@{}c@{}}MPJPE\\ (mm)\end{tabular} \\ \cline{1-4}
dot                & fcl               & most-conf    & all   &       \\ \hline
                  &                   &                 &       & 22.60  \\ \hline
    \checkmark               &                   &  \checkmark                   &     & 22.05  \\
      \checkmark                 &               &                   &  \checkmark         & 21.88  \\ \hline
                   &  \checkmark                     &    \checkmark                   &       & 21.92  \\
                   &   \checkmark                    &                   &   \checkmark        & \textbf{21.61}  \\ \Xhline{0.09em}
\end{tabular}}
\end{center}
\end{minipage}
\hspace{1ex}
\begin{minipage}{0.25\linewidth}
\begin{center}
\caption{\textbf{Effect of loss strategy.} PJ: reprojected loss, BL: bone length loss, JA: joint angle loss.}
\label{tb:exp_loss_strategy}
\resizebox{\textwidth}{!}{
\begin{tabular}{ccc|c|c}
\Xhline{0.09em}
\multicolumn{3}{c|}{Strategy}           & \multirow{2}{*}{\begin{tabular}[c]{@{}c@{}}MPJPE\\ (mm)\end{tabular}} & \multirow{2}{*}{\begin{tabular}[c]{@{}c@{}}PPP$@0.2$\\ (\%)\end{tabular}} \\ \cline{1-3}
PJ   &BL       &JA  &                                &                                              \\ \hline
         &          &          & 22.60                           & 79.36                           \\ \hline
\checkmark &        &              & 22.12                          & 79.37                         \\
&\checkmark        &              & 22.23                          & 79.55                         \\
          & & \checkmark          & 22.40                           & 79.43                \\ \hline
 &\checkmark        & \checkmark          & 22.10                           & 79.68                         \\
\checkmark        & \checkmark  & \checkmark                    & 21.89                          & 79.70                      \\ \Xhline{0.09em}
\end{tabular}}
\end{center}
\end{minipage}
\hspace{1ex}   %%两个minipage之间相隔3个字符的距离
\begin{minipage}{0.4\linewidth}
\caption{\textbf{Effect of the number of views during testing in terms of MPJPE.} When combined with MVF, the case of using three views is not tested because it takes a long time to train.}
\vspace{1.7em}
\label{tb:exp_number_of_views}
\resizebox{\textwidth}{!}{
\begin{tabular}{c|cccc}
\Xhline{0.09em}
\multirow{2}{*}{$\#$(views)} & \multicolumn{4}{c}{MPJPE (mm)} \\ \cline{2-5} 
                                     & baseline        & ours-HT        & ours-MVF       & ours       \\ \hline
4                                    & 22.60            & 21.58        & 21.61          & 21.12      \\ 
3                                    & 27.08           & 26.11        & -          & -       \\ 
2                                    & 33.43           & 31.83        & 30.74          & 29.99       \\ \Xhline{0.09em}
\end{tabular}}
\end{minipage}
\vspace{-1em}
\end{table*}

\noindent\textbf{Qualitative Results on Human3.6M.}
To evaluate the multi-view consistency performance of MVF, the estimated, reprojected and groundtruth 2D keypoints are compared. The estimated 2D keypoint will be close to the reprojected keypoint from 3D result if the keypoint is consistent with other views. As illustrated in Fig.~\ref{fig:exp_MVFvisualization}, the blue (reprojected) and green (estimated) points are generally closer after MVF refinement, especially for some self-occlusion. The improvement suggests that the MVF module can make views perceive others and provide assistant for some unseen view from other seen views. 

Furthermore, qualitative experiments are used to evaluate the ability to amend the implausible pose of HT approach. As illustrated in Fig.~\ref{fig:exp_qualitative_results}, the extracted anatomy prior can amend some unreasonable errors. It is particularly noteworthy that HT has the capability to correct the pose to have normal joint angles and bone lengths in such a tough situation.

\noindent\textbf{Generalization to the Total Capture Dataset.}
To substantiate generalization of our model, we first conduct cross-dataset experiments on Total Capture, the testing model is only trained by Human3.6M training set. As shown in Table~\ref{tb:exp_compare_with_sota_tc}, our method surpasses baseline by $12mm$ ($19\%$), which demonstrates the generalization of our method.
% And we observe that the errors on each action are similar. Hence, it can be verified that the anatomy coherence extracted has the capability to abstract pose regardless of the diversity of motions. 
Then we train our model with Total Capture training data under the same strategy clarified in Sec.\ref{sec:implementation_detail}. Our method achieves $23mm$ MPJPE, which also exceeds GeoFuse~\citep{zhang2020fusing} by 8\%. It is worth noting that, the PCA training set does not contain the Total Capture, which demonstrates that our anatomy coherence has the ability to deal with unseen gestures.

\subsection{Ablation Study\label{sec:exp_ablation_study}}
All ablation studies are conducted on the Human3.6M. Both MPJPE and PPP$@0.2$ metrics are used for evaluation. 

\noindent\textbf{MVF Module Evaluation.}
Beside the matching strategy, we also evaluate the number of views when fusing, there are two pipelines: (1) all view fusion, each view generates pseudo heatmap assist to other views, (2) most-conf fusion, only most-confident view is chosen to generate pseudo heatmap. The comparison is shown in Table~\ref{tb:exp_matching_strategy}, fully connected layer slightly outperforms the inner dot matching. And all view fusion performs better. We conjecture that since most initial results are reliable, as the number of fused views increases, more accurate auxiliary information is provided.

\noindent\textbf{Effect of Loss.}
As shown in Table~\ref{tb:exp_loss_strategy}, PJ loss brings $2.1\%$ relative MPJPE improvement, which is more than bone length loss and joint angle loss. We suppose it is because that view consistency enhance correspondence of multi-view 2D keypoints. But the plausibility raises little with PJ loss. BL loss and JA loss bring more improvement in PPP$@0.2$. It demonstrates that the kinematic skeleton structure can boost the plausibility of pose. Finally we retrain the network which combines all loss function, and obtain the results whose MPJPE is $21.89mm$ and PPP$@0.2$ is $79.7\%$.

\noindent\textbf{Effect of the Number of Views.}
Views are reduced from 4 to 2 during testing to explore influence of the number of views.
% and evaluate generalization to different strength of geometric constraints. 
As the number of views decreases, precision degrades. But MPJPE equals to $29.99mm$ when there are two views as shown in Table~\ref{tb:exp_number_of_views}, which is still excellent. And compare ours with ours-HT or ours-MVF, the conclusion that both view consistency and anatomy coherence can improve pose estimation is proved.
% As illustrated in Fig.~\ref{fig:exp_number_of_views}, the improvement gap of plausibility also increases as the number of views decreases, which demonstrates that our anatomy prior is complementary to the reduction of geometric constraints. 

\noindent\textbf{Effect of Orientation Normalization.}
Whether and what the anatomy coherence PCA factors out still bother us. To observe it, we change each latent variable individually with a small step to generate the recovered 3D pose. The changes in 3D poses represents the physical meaning of the corresponding latent variable. Without orientation normalization, there are only 8 out of 25 latent variables describe joint correlations in motion, and the remaining 17 describe rotation invariant property. And the results have been improved after orientation normalization. The 25 variables all describe the joint-coupled motion. And the transformation of the recovered 3D poses demonstrate the capability of PCA to restrict joints correlation through motion. 
 % Hence, although orientation normalization brings little performance improvement, we still adopt this preprocessing.

\section{Conclusion}\label{}
We propose view consistency aware holistic triangulation to improve the performance of both precision and plausibility in 3D HPE. The key contribution is that the geometric correspondences of multi-view 2D keypoints are enhanced and anatomy coherence is injected to 2D-3D process. Meanwhile, a PPP metric is raised to evaluate the pose plausibility. Experiments not only exhibit that our approach outperforms state-of-the-art methods, but also demonstrate that the reconstruction term with extracted skeletal structure features can abstract the human anatomy prior.
 % For further work, more effort is hope to made on the metric to get rid of the data dependency and empirical hyperparameter, 2D immediate features are expected to be fused with triangulation, anatomy prior extraction and aggregation is desirable to be further explored.

% \section{References}
\bibliographystyle{model2-names}
\bibliography{refs}

\clearpage
\twocolumn[\begin{@twocolumnfalse}
\Large
\centerline{
\rule[5pt]{0.05\textwidth}{0.05em}Supplementary Materials\rule[5pt]{0.05\textwidth}{0.05em}
}
View Consistency Aware Holistic Triangulation for 3D Human Pose Estimation
\normalsize
\vspace{5em}
\end{@twocolumnfalse}]

\setcounter{section}{0}
\setcounter{table}{0}
\setcounter{figure}{0}
\section{2D Backbone Structure}
The 2D backbone consists of ResNet-152 and three deconvolutional layers to produce the immediate features, followed by a $1\times1$ convolutional kernel to generate 2D heatmaps $H_{(c,k)}$ with $K$ output channels. Where $k\in \{1,2,\cdots,K\}$, $K$ is the number of keypoints, and $c$ is the index of view. 2D positions are calculated by soft-argmax, a differentiable approach which makes end-to-end training of 3D estimator possible. Soft-argmax treats the 2D keypoints $\mathbf{x}_k$ as the centroid of heatmap $H_{k}$, and the weight of each pixel is produced by softmax in heatmap:
\begin{gather*}
    \mathbf{x}_k = \sum\limits_{r_x}^{W} \sum\limits_{r_y}^{H} 
    \mathbf{r} \cdot H_{k}^{'}(\mathbf{r}) \\
    H_{k}^{'}(\mathbf{r}) = \frac{e^{H_{k}(\mathbf{r})}}{\sum_{\mathbf{r}} e^{H_{k}(\mathbf{r})}}
\label{eq:supple_2d_loc}
\end{gather*}
where $W,H$ are width and height of the heatmap respectively, and $\mathbf{r}$ is the location of the pixel in heatmap.

\setlength{\tabcolsep}{2pt}
\begin{table*}[t]
% \begin{minipage}{\linewidth}
\begin{center}
\caption{\textbf{Low dimension preserved design comparison.} We refer to baseline as $D=51$ whose dimension is not reduced.}
\label{tb:exp_low_dimension_pca}
\vspace{0.5em}
\resizebox{0.5\textwidth}{!}{
\begin{tabular}{lccccccc}
\toprule
Dimension $D$ & 51     & 35    & 30    & \textbf{25}    & 20    & 15    & 10    \\ \midrule
MPJPE- (mm) & 22.60 & 22.10 & 22.04  &\textbf{22.04}  & 22.06 & 22.08 & 22.11 \\ 
PPP@0.2 (\%)  & 79.36 & 79.90 &79.91  & \textbf{80.97} & 80.14 & 80.19 & 80.33  \\ \bottomrule
\end{tabular}}
\vspace{-1.5em}
\end{center}
% \end{minipage}

% \vspace{2ex} 

% \begin{minipage}{\linewidth}
% \begin{center}
% \caption{Reconstruction term coefficient comparison. The magnitude of $\lambda$ is $10^3$.}
% \label{tb:exp_coefficient_pca}
% \begin{tabular}{l|cccccccccc}
% \hline
% Coefficient $\lambda$ & 1     & 2     & 3     & 4     & 5     & 6     & 7     & \textbf{8}     & 9     & 10    \\ \hline
% MPJPE-re (mm) & 22.11 & 22.09 & 22.07 & 22.06 & 22.05 & 22.04 & 22.04 & \textbf{22.04} & 22.05 & 22.06 \\ 
% PRP@0.2 (\%)  & 79.68 & 79.68 & 80.19 & 80.55 & 80.60 & 80.55 & 80.74 & \textbf{80.97} & 81.02 & 80.74 \\ \hline
% \end{tabular}
% \end{center}
% \end{minipage}
\end{table*}

\setlength{\tabcolsep}{1.4pt}

% \setlength{\tabcolsep}{5pt}
% \begin{table}[h]
% \begin{center}
% \caption{Low dimension preserved design comparison. We refer to baseline as $D=0$.}
% \label{tb:exp_low_dimension_pca}
% \begin{tabular}{l|c|c|c|c|c|c|c}
% \hline
% Dimension $D$ & 0     & 10    & 15    & 20    & \textbf{25}    & 30    & 35    \\ \hline
% MPJPE-re (mm) & 22.60 & 22.11 & 22.08 & 22.06 & \textbf{22.04} & 22.04 & 22.10 \\ \hline
% PRP@0.2 (\%)  & 79.36 & 80.33 & 80.19 & 80.14 & \textbf{80.97} & 79.91 & 79.90 \\ \hline
% \end{tabular}
% \end{center}
% \end{table}
% \setlength{\tabcolsep}{1.4pt}

% \setlength{\tabcolsep}{2.5pt}
% \begin{table}[b]
% \begin{center}
% \caption{Reconstruction term coefficient comparison. The magnitude of $\lambda$ is $10^3$.}
% \label{tb:exp_coefficient_pca}
% \begin{tabular}{l|c|c|c|c|c|c|c|c|c|c}
% \hline
% Coefficient $\lambda$ & 1     & 2     & 3     & 4     & 5     & 6     & 7     & \textbf{8}     & 9     & 10    \\ \hline
% MPJPE-re (mm) & 22.11 & 22.09 & 22.07 & 22.06 & 22.05 & 22.04 & 22.04 & \textbf{22.04} & 22.05 & 22.06 \\ \hline
% PRP@0.2 (\%)  & 79.68 & 79.68 & 80.19 & 80.55 & 80.60 & 80.55 & 80.74 & \textbf{80.97} & 81.02 & 80.74 \\ \hline
% \end{tabular}
% \end{center}
% \end{table}
% \setlength{\tabcolsep}{1.4pt}
\begin{table*}[t]
\begin{center}
\caption{\textbf{Reconstruction term coefficient comparison.}}
\label{tb:exp_coefficient_pca}
\setlength{\tabcolsep}{5pt}
\vspace{0.5em}
\resizebox{0.75\textwidth}{!}{
\begin{tabular}{l|cccccccccc}
\Xhline{0.09em}
Coefficient $\lambda$                                            & 1     & 2     & 3     & 4     & 5   & 6     & 7     & \textbf{8}     & 9     & 10  \\ \Xhline{0.05em}
MPJPE (mm) & 22.11 & 22.09 & 22.07 & 22.06 & 22.05 & 22.04 & 22.04 & \textbf{22.04} & 22.05 & 22.06\\
PPP@0.2 (\%)  & 79.68 & 79.68 & 80.19 & 80.55 & 80.60 & 80.55 & 80.74 & \textbf{80.97} & 81.02 & 80.74\\
\Xhline{0.09em}
\end{tabular}}
\end{center}
\end{table*}
\setlength{\tabcolsep}{2pt}
\begin{table*}[!t]
% \begin{minipage}{\linewidth}
\begin{center}
\caption{Coefficient comparison of hop$\_1$ feature. The magnitude of $\lambda$ is $10^{3}$.}
\label{tb:exp_low_dimension_dis1}
\begin{tabular}{lcccccccccc}
\toprule
Coefficient $\lambda$ & 1     & 2    & 3    & \textbf{4}    & 5    & 6   & 7   & 8    & 9    & 10  \\ \midrule
MPJPE-re (mm) & 22.03 & 21.94 & 21.88 & \textbf{21.86} & 21.86 & 21.90 & 21.96 & 22.04 & 22.15  & 22.27 \\ 
PRP@0.2 (\%)  & 79.55 & 80.01 & 80.37 & \textbf{80.92} & 80.65 & 80.92 & 80.92  & 80.88  & 81.11  & 81.20 \\ \bottomrule
\end{tabular}
\end{center}
\end{table*}
\setlength{\tabcolsep}{2pt}
\begin{table*}[!t]
% \begin{minipage}{\linewidth}
\begin{center}
\caption{Coefficient comparison of hop$\_2$ feature. The magnitude of $\lambda$ is $10^{3}$.}
\label{tb:exp_low_dimension_dis2}
\begin{tabular}{lcccccccccc}
\toprule
Coefficient $\lambda$ & 1     & 2    & 3    & \textbf{4}    & 5    & 6   & 7   & 8    & 9    & 10  \\ \midrule
MPJPE-re (mm) & 22.06 & 21.98 & 21.94 & \textbf{21.91} & 21.91 & 21.92 & 21.95 & 21.99 & 22.05  & 22.11 \\ 
PRP@0.2 (\%)  & 79.87 & 80.15 & 80.56 & \textbf{80.79} & 80.65 & 80.33 & 80.55  & 81.02  & 81.24  & 81.15 \\ \bottomrule
\end{tabular}
\end{center}
\end{table*}

\section{Linear Triangulation Theory}
The homogeneous 2D $\widetilde{\mathbf{x}}_{c,k}$ and 3D $\widetilde{\mathbf{y}}_{k}$ keypoint can be expressed as $\widetilde{\mathbf{x}}_{c,k} = w_{c,k}[u_{c,k}, v_{c,k}, 1]^{T}$, $\widetilde{\mathbf{y}}_{k} = [x, y, z, 1]^{T}$. The projected 2D keypoint from $\widetilde{\mathbf{y}}_{k}$ in the $c^{th}$ view is $P_c\widetilde{\mathbf{y}}_{k}$. To make the estimated  $\widetilde{\mathbf{x}}_{c,k}$ and projected $P_c\widetilde{\mathbf{y}}_{k}$ same, the target function can be written as:
\begin{gather*}
    \widetilde{\mathbf{x}}_{c,k} = P_c\widetilde{\mathbf{y}}_{k}
\label{eq:supple_linear_tri_1}
\end{gather*}
and then:
\begin{gather*}
    w_{c,k}u_{c,k} = P_c^{1}\widetilde{\mathbf{y}}_{k},\quad w_{c,k}v_{c,k} = P_c^{2}\widetilde{\mathbf{y}}_{k},\quad w_{c,k} = P_c^{3}\widetilde{\mathbf{y}}_{k}
\label{eq:supple_linear_tri_2}
\end{gather*}
where $P_c^{i}$ is the $i^{th}$ row of $P_c$, so:
\begin{gather*}
    u_{c,k}P_c^{3}\widetilde{\mathbf{y}}_{k} -  P_c^{1}\widetilde{\mathbf{y}}_{k} = 0 \\
    v_{c,k}P_c^{3}\widetilde{\mathbf{y}}_{k} -
    P_c^{2}\widetilde{\mathbf{y}}_{k} = 0
\label{eq:supple_linear_tri_3}
\end{gather*}

Finally, the target function can be expressed as $\widetilde{A}_k\widetilde{\mathbf{y}}_{k}=0$, where 
\begin{gather*}
    \widetilde{A}_k = 
    \begin{bmatrix}
    & u_{1,k}P_1^{3} - P_1^{1}\\
    & v_{1,k}P_1^{3} - P_1^{2}\\
    & \vdots \\
    & u_{C,k}P_C^{3} - P_C^{1}\\
    & v_{C,k}P_C^{3} - P_C^{2}
    \end{bmatrix}
\label{eq:supple_linear_tri_4}
\end{gather*}

Change the 3D keypoint to inhomogeneous expression $\mathbf{y}_{k}$, the target function can be rewritten like:
\begin{gather*}
    A_k \mathbf{y}_k + \mathbf{b}_k = 0
\label{eq:supple_linear_tri_4}
\end{gather*}
where $A_k$ is the first three columns of $\widetilde{A}_k$ and $\mathbf{b}_k$ is the last column.

\section{The Closed-Form Solution Proof}
The objective function of Holistic Triangulation with anatomy constraints is:
\begin{gather*}
    \min\limits_{Y} \quad  \Vert A Y + B\Vert + \lambda \Vert Y_{re} - Y_{re}^{'} \Vert
\label{eq:supple_holistic_tri_1}
\end{gather*}
where $Y_{re}^{'} = M^{T}M(Y_{re}-Y_{mean})+Y_{mean}$ is the recovered pose; $M\in\mathbb{R}^{(D,3K)}$ is the feature extracting matrix of PCA encoder; $Y_{mean}\in\mathbb{R}^{(3K,1)}$ is the mean pose of PCA training set; $Y_{re}=Y-Y_{root}$; and $\lambda$ is a learnable weight of reconstruction term. Then objective function can be rewritten as:
\begin{small}
\begin{gather*}
    \min\limits_{Y} \quad  \Vert A Y + B\Vert + \lambda \Vert (I-M^{T}M)Y-(I-M^{T}M)(Y_{root}+Y_{mean}) \Vert
\label{eq:supple_holistic_tri_2}
\end{gather*}
\end{small}

Because of the property preserving convexity that nonnegative weighted sums and composition with an affine mapping have, the function is still convex. So, using Least-Square method to solve it, the closed form answer can be expressed as:
\begin{gather*}
    2(A^{T}AY + A^{T}B) + \\
    2[\lambda(I-M^{T}M)^{T}(I-M^{T}M)Y \\
    - \lambda(I-M^{T}M)^{T}(I-M^{T}M)(Y_{root}+Y_{mean})] \\
    = 0
\label{eq:supple_holistic_tri_3}
\end{gather*}

So the answer can be simplified to:
\begin{gather*}
    (A^{T}A+\lambda N^{T}N)Y = \lambda N^{T}N(Y_{root}+Y_{mean})-A^{T}B 
\label{eq:supple_holistic_tri_4}
\end{gather*}
where $N = (I-M^{T}M)$.

\section{KCS Details}
The original KCS is a algorithm to create a vector between  two defined joints using matrix multiplication. The target vector from $\mathbf{y}_r$ to $\mathbf{y}_l$ can be calculated as:
\begin{gather*}
    \mathbf{b}_j = \mathbf{y}_l - \mathbf{y}_r = \widetilde{Y}c_j
\label{eq:supple_kcs_1}
\end{gather*}
where $\mathbf{c}_j=[0,\cdots,0,1,0,\cdots,0,-1,0,\cdots,0]^{T}\in \mathbb{R}^{(K,1)}$ and $\widetilde{Y} \in \mathbb{R}^{(3,K)}$ is the matrix of all 3D keypoints. To compatible with our 3D keypoints expression $Y \in \mathbb{R}^{(3K,1)}$, the KCS can be rewritten as:
\begin{gather*}
V = CY 
\label{eq:supple_kcs_2}\\
Y = GV
\label{eq:method_kcs_verse}
\end{gather*}
where $C=[\mathbf{c}_{1}^{T},\mathbf{c}_{2}^{T},\cdots,\mathbf{c}_{J}^{T}]^{T} \in \mathbb{R}^{(3J,3K)}$ is the matrix which maps the keypoint location vector $Y$ to bone vector $V$; $G \in \mathbb{R}^{(3K,3J)}$ is the matrix maps $V$ back to $Y$; and
\begin{gather*}
\mathbf{c}_j=
\left[ \begin{array}{@{}*{11}{c}@{}}
0 &\cdots &1 &0 &0 &\cdots &-1 &0 &0 &\cdots &0 \\
0 &\cdots &0 &1 &0 &\cdots &0 &-1 &0 &\cdots &0 \\
0 &\cdots &0 &0 &1 &\cdots &0 &0 &-1 &\cdots &0
\end{array} \right] \in \mathbb{R}^{(3,3K)}
\label{eq:supple_kcs_3}
\end{gather*}
which can transform two keypoints with index $l,r$ to the connected bone vector by $\mathbf{y}_{l}-\mathbf{y}_{r}$.

\begin{figure}[t]
\begin{minipage}[t]{\linewidth}
\centering
\includegraphics[width=0.8\textwidth]{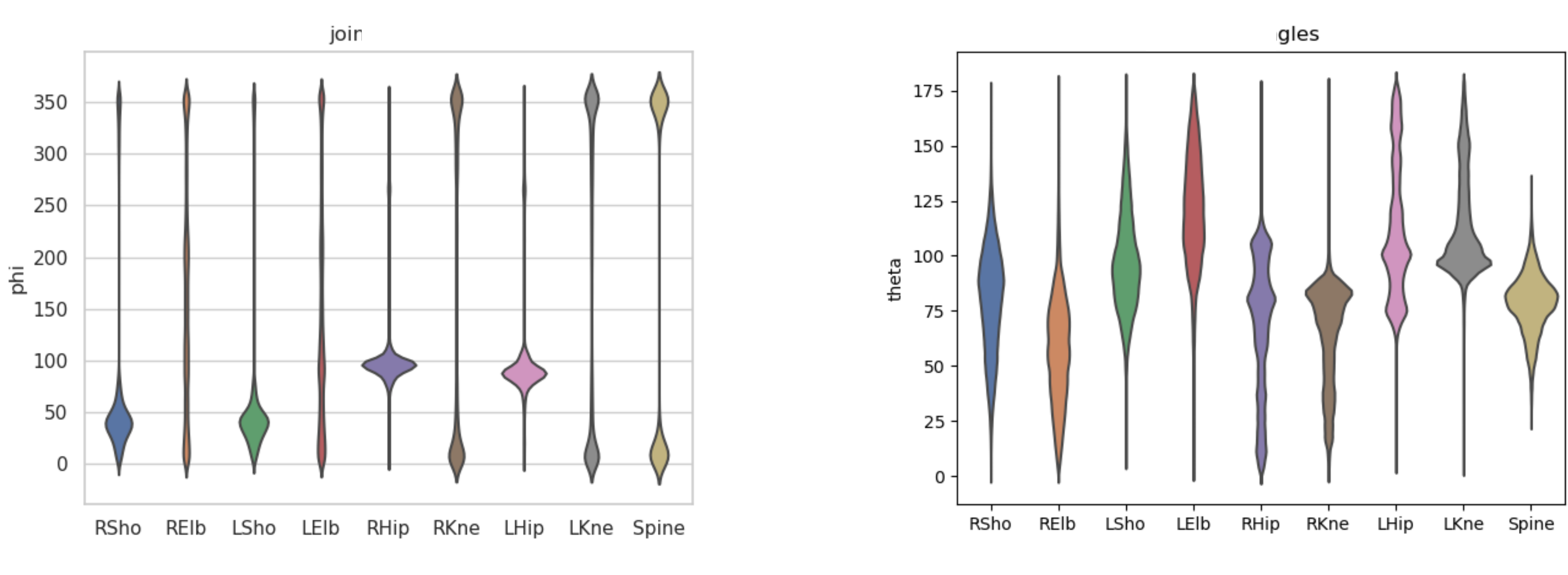}
\caption{Illustration of 2D distribution of selected joints' angles, the left image is the distribution of azimuth angle and the right one is polar angle.}
\label{fig:method_joint_angle_distribution_2D}
\end{minipage}
\begin{minipage}[]{\linewidth}
\centering
\includegraphics[width=0.8\textwidth]{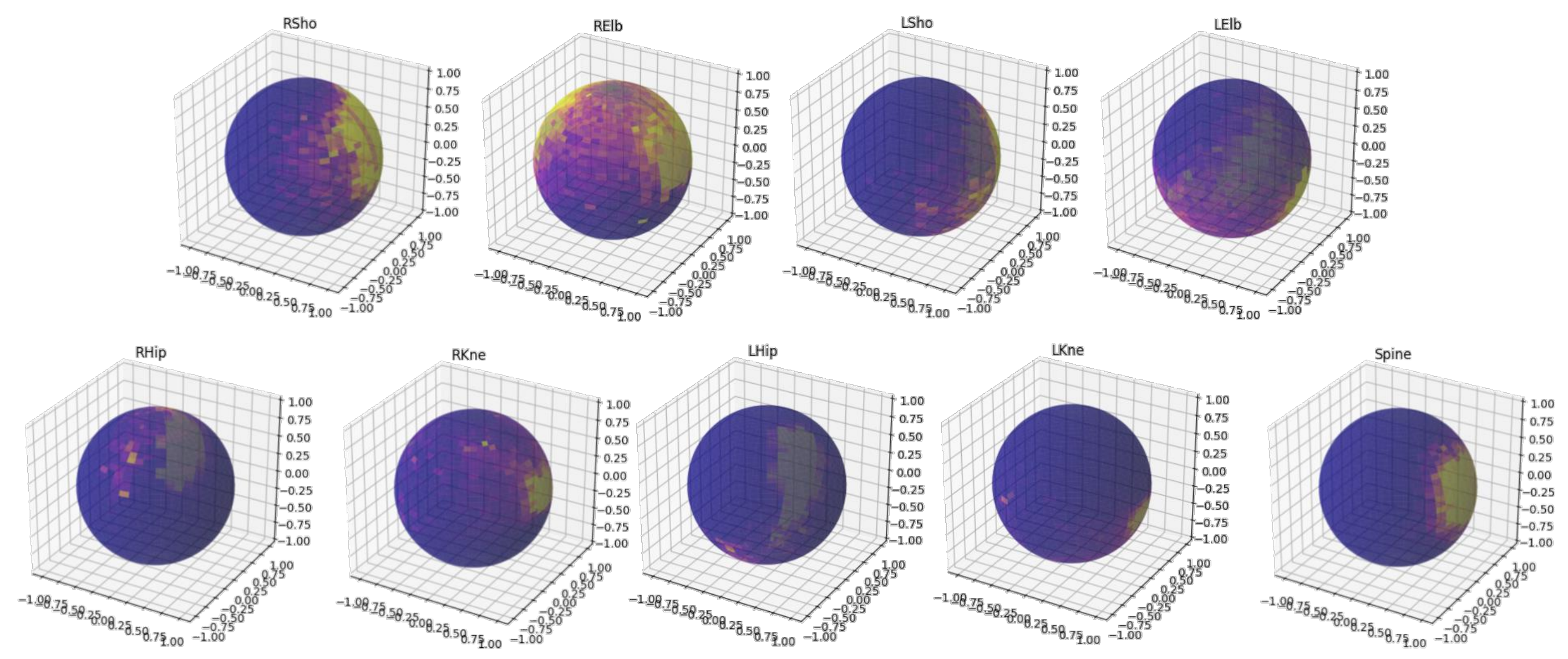}
\caption{Illustration of 3D distribution of selected joints' angles.}
\label{fig:method_joint_angle_distribution_3D}
\end{minipage}
\end{figure}

% \begin{figure}[!t]
% \centering
% \includegraphics[width=0.9\textwidth]{experiment/tc_definition_difference.pdf}
% \caption{Illustration of definition differences between Total Capture and Human3.6M. The first row is the groundtruth of Total Capture and the second row is the projected 2D pose from 3D pose estimated by our HT method.}
% \label{fig:exp_definition_difference_between_datasets}
% \end{figure}

\section{Loss Detail}
\noindent\textbf{Reprojected Loss.}
The reprojected loss is expressed as the sum of the errors between reprojected keypoints from $Y$ and refined keypoints $X^{'}$:
% \vspace{-0.1em}
\begin{equation*}
    L_{pj} = \frac{1}{C} \frac{1}{K} \sum\limits_{c=0}^{C}\sum\limits_{k=0}^{K} \Vert \mathbf{x}^{'}_{c,k}- P_{c}\mathbf{y}_{k} \Vert
\label{eq:method_pj_loss}
\end{equation*}

\noindent\textbf{Bone Length Loss.}
L1 norm is utilized to measure distance of bone length between the solved pose and the annotation:
\vspace{-0.1em}
\begin{equation*}
    L_{bl} = \frac{1}{J}\sum\limits_{j=0}^{J} \Vert BL_{j}- \hat{BL}_{j} \Vert_1
\label{eq:method_bl_loss}
\end{equation*}
where $BL_{j}$ represents the $j^{th}$ bone length and $\hat{BL_{j}}$ is the groundtruth.
% ; $BL_{j}$ is the modulo of the bone vector calculated by KCS.

\noindent\textbf{Joint Angle Distribution}
The angle of the $k^{th}$ joint is defined as the angle between two bones connected to this joint. To calculate it, we convert the global coordinate to a local spherical coordinate system and get the azimuth $\varphi_{k}$ and polar angle $\theta_{k}$.

The distributions of all selected angles almost fit the Gaussian Mixture Module (GMM), which can be illustrated as in Fig.~\ref{fig:method_joint_angle_distribution_2D}. After expanding into 3D space, the distribution can be observed more intuitively in Fig.~\ref{fig:method_joint_angle_distribution_3D}. Here, we select nine joints angles: right-shoulder, right-elbow, left-shoulder, left-elbow, right-hip, right-knee, left-hip, left-knee, and spine.
The multivariate GMM is used to model the distribution of joint angle in this way:
% which is illustrated in the supplementary. 
% Therefore, we apply the multivariate GMM to model the distributions of joints angles:
\begin{equation*}
    p(x_k \vert \Theta) = \sum\limits_{i}\alpha_{i}\phi(x_k\vert \Theta_i)
\label{eq:method_multivariate_GMM}
\end{equation*}
where $\Theta_i$ and $\alpha_{i}$ are parameters and weights of the module; the angle $(\varphi_{k},\theta_{k})$ of the $k^{th}$ joint is represent by $x_k=[\sin{\theta_k}, \sin{\varphi_k}, \cos{\varphi_k}]^{T}$. Trigonometric functions are used to address the discontinuity stepping from $360^{\circ}$ to $0^{\circ}$.
% where $\Theta_i$ and $\alpha_{i}$ are parameters and coefficients of the module with index $i$, and $x_k$ represents the angle $(\varphi_{k},\theta_{k})$ of the $k^{th}$ joint. To address the discontinuity when stepping from $360^{\circ}$ to $0^{\circ}$, we convert angles using trigonometric functions, therefore $x_k=[\sin{\theta_k}, \sin{\varphi_k}, \cos{\varphi_k}]^{T}$.

% \section{Definition Difference between Total Capture and Human3.6M}
% Differences of definition between Total Capture and Human3.6M are illustrated in Fig.~\ref{fig:exp_definition_difference_between_datasets}. The whole pose of Total Capture leans to the right, two ``hip" points are closer than the condition in Human3.6M, ``head" point is also lower than Human3.6Ms, and ``spine" point of Total Capture is on the body surface while Human3.6Ms is on the bone. It should be mentioned that the point of ``neck" in Human3.6M can not match any labels in Total Capture, so we fix this point to $(0,0,0)$ and will not evaluate it statistically. 

\section{HT Module Design}
It should be mentioned that HT is only used as a post-processing step, replacing the baseline triangulation, to determine the hyperparameters.

First, we compare different choices of low-dimension $D$ preserved by PCA. As the dimension decreases, less variance is kept, which means lower precision but higher abstraction. As shown in Table~\ref{tb:exp_low_dimension_pca}, we change $D$ from $35$ to $10$ with stride $5$, which corresponds to preserving variance from $99.9\%$ to $88.3\%$. As $D$ decreases, both precision MPJPE and PPP get improved, and peaks at $D=25$ ($99.5\%$ variances) where a balance is struck in prior extraction and precision preservation. Compared with baseline, the improvement demonstrates the importance of PCA reconstruction term, implying that the pose global context is extracted.

\begin{figure}[t]

\begin{minipage}[t]{\linewidth}
    \centering
    \includegraphics[width=\textwidth]{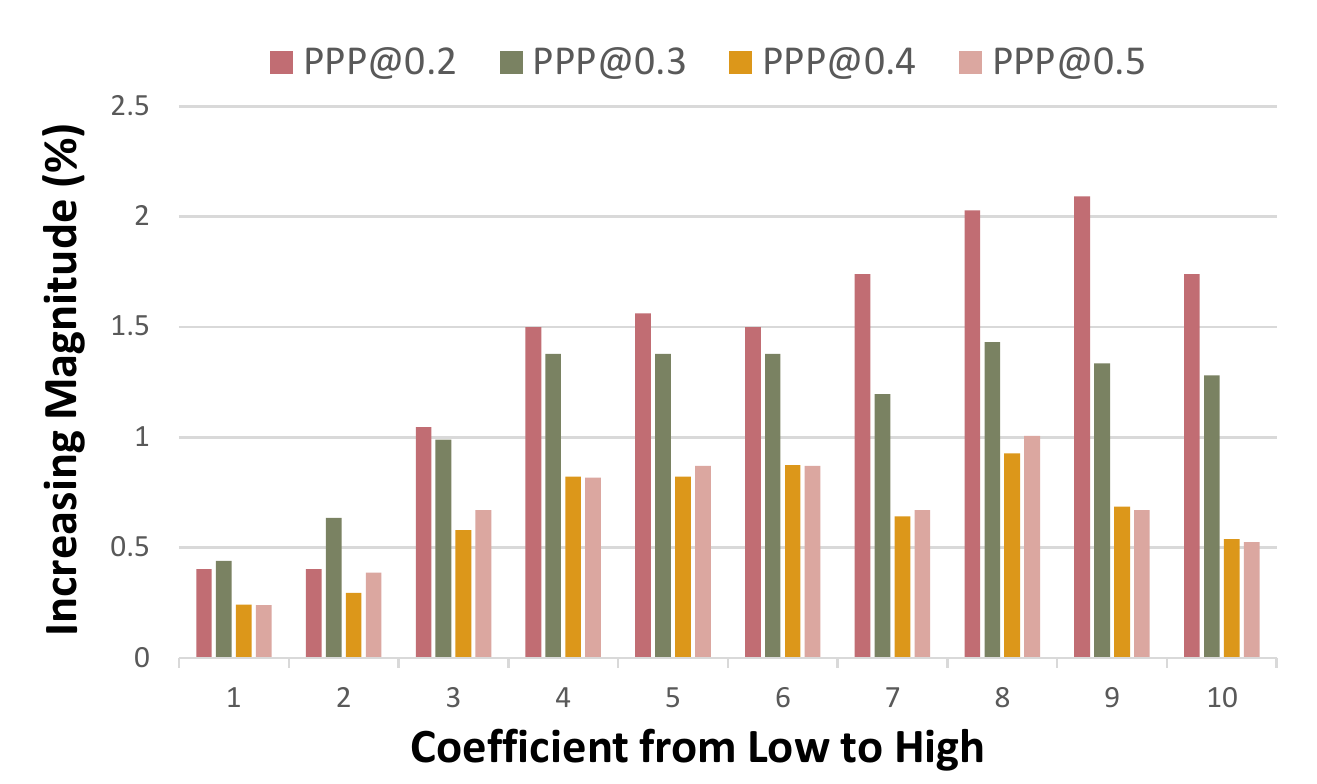}
    % \vspace{-2em}
    \caption{The increasing magnitude compared to baseline of PPP$@R$ with \textbf{various coefficient.}}
    % Different colors represent $R=0.2,0.3,0.4,0.5$ respectively.}
    \label{fig:exp_coefficient_effect_PRP}
\end{minipage}
% \end{figure}
% \begin{figure}[t]
% \hspace{0.5em}   %%两个minipage之间相隔3个字符的距离
% \begin{minipage}[t]{\linewidth}
%     \centering
%     \includegraphics[width=\textwidth]{experiment/number_of_views.pdf}
%     \vspace{-2em}
%     \caption{The comparison of baseline and ours-HT in terms of PPP$@0.2$ with \textbf{different numbers of views} during testing.}
%     \label{fig:exp_number_of_views}
% \end{minipage}
% \vspace{-1em}
\end{figure}

Then, different coefficients $\lambda$ of vanilla reconstruction term ($hop\_0$ feature, others can be found in supplementary) are evaluated. The magnitude of $\lambda$ is set to $10^3$ to make the reprojection term and reconstruction term approximate $10:1$. $\lambda=8$ is our final choice due to its balance between MPJPE and PPP$@0.2$, as shown in Table~\ref{tb:exp_coefficient_pca}. The increase of $\lambda$ does not bring much improvement to MPJPE but improves more on PPP$@0.2$. For one thing, the coefficient variation has little effect on the accuracy when two terms are of the same magnitude. For another, the observation that average accuracy does not equal to reasonability can be verified. We compare performance on different threshold $R$ of PPP which represent different tolerance for implausible pose. As illustrated in Fig.~\ref{fig:exp_coefficient_effect_PRP}, reconstruction term almost yields more than $0.5\%$ growth compared with baseline regardless of different $R$ and $\lambda$. It demonstrates that our anatomy term can improve the poses with different degrees of implausibility, especially when the degree is low.

We finally use training set to extract different $hop$ vectors, and then generate corresponding PCA extraction matrices $M$ to compare different feature extraction strategy. As shown in Table~\ref{tb:exp_structure_strategy}, $hop\_1$ which conforms to the human skeleton performs slightly better, and precision in $hop\_2$ also surpasses $hop\_0$, which can justify the effect of skeleton structural feature. Then, we fuse them, the strategy which combines three types yields improvement by $0.82mm$ (relative $3.6\%$) in MPJPE and $1.88\%$ (relative $2.4\%$) in PPP$@0.2$. It should be noted that, adding different features does not bring corresponding increase as same as the individual condition. We hypothesize that all features are extracted from one data source, so stacking them does not yield much information gain.
\begin{table}[t]{}
\begin{center}
\caption{\textbf{Effect of skeletal structural feature strategies.} First row is baseline.}
\label{tb:exp_structure_strategy}

\begin{tabular}{ccc|c|c}
\Xhline{0.09em}
\multicolumn{3}{c|}{Strategy} & \multirow{2}{*}{\begin{tabular}[c]{@{}c@{}}MPJPE\\ (mm)\end{tabular}} & \multirow{2}{*}{\begin{tabular}[c]{@{}c@{}}PPP$@0.2$\\ ($\%$)\end{tabular}} \\ \cline{1-3}
hop=0    & hop=1    & hop=2   &                                &                                 \\ \hline
         &          &         & 22.60                          & 79.36                           \\ \hline
\checkmark        &          &         & 22.04                          & 80.97                           \\
         & \checkmark        &         & 21.86                          & 80.92                           \\
         &          & \checkmark       & 21.91                          & 80.79                           \\ \hline
\checkmark        & \checkmark        &         & 21.83                          & 80.97                           \\
\checkmark        & \checkmark        & \checkmark       & \textbf{21.78}                          & \textbf{81.24}                           \\ \Xhline{0.09em}
\end{tabular}
\end{center}
\end{table}

The low dimension preserved by PCA of hop$\_1$ and hop$\_2$ are set to $20$ and $15$ to keep the similar variances with hop$\_0$. And the coefficients comparison are shown in Table~\ref{tb:exp_low_dimension_dis1} and Table~\ref{tb:exp_low_dimension_dis2}. Finally, the initial coefficient of hop$\_1$ and hop$\_2$ are set to $4000$, $4000$ because of the balance between precision and plausibility.

% \section{MVF Module Design}
% Beside the matching strategy, we also evaluate the number of views when fusing, there are two pipelines: (1) all view fusion, each view generates pseudo heatmap assist to other views, (2) most-conf fusion, only most-confident view is chosen to generate pseudo heatmap. The comparison is shown in Table~\ref{tb:exp_matching_strategy}, fully connected layer slightly outperforms the inner dot matching. And all view fusion performs better. We conjecture that since most initial results are reliable, as the number of fused views increases, more accurate auxiliary information is provided.






\end{document}